\documentclass[]{xiaomiev}

\usepackage[toc,page,header]{appendix}
\usepackage{booktabs}
\usepackage{multirow}
\usepackage{graphicx}
\usepackage{array}
\usepackage{colortbl}
\usepackage{needspace}
\usepackage{makecell}
\usepackage{amsmath}
\usepackage{amssymb}
\usepackage{placeins}
\usepackage{longtable}
\usepackage{float}
\usepackage{cuted}
\usepackage{chngcntr}
\usepackage{cleveref}
\usepackage{threeparttable}
\usepackage{tabularx}
\usepackage{subcaption}
\usepackage{wrapfig}
\usepackage{enumitem}
\usepackage{pifont}
\usepackage{adjustbox}
\usetikzlibrary{arrows.meta,positioning,fit,calc,backgrounds}
\definecolor{tableorange}{RGB}{250,225,207}
\setlist[itemize]{leftmargin=15pt}
\usepackage{flafter}

\AddToHook{cmd/section/before}{\par\Needspace{5\baselineskip}}
\titlespacing*{\section}{0pt}{14pt plus 3pt minus 2pt}{7pt plus 1pt minus 1pt}
\titlespacing*{\subsection}{0pt}{11pt plus 2pt minus 1pt}{5pt plus 1pt minus 1pt}
\titlespacing*{\subsubsection}{0pt}{8pt plus 2pt minus 1pt}{4pt plus 1pt minus 1pt}

\renewcommand{\paragraph}[1]{\noindent\textbf{#1.}\hspace*{1em}}
\newcommand{\method}{OmniBrain-0}
\newcommand{\collabicon}{\raisebox{0.08ex}{\textcolor{xiaomievblue}{\scalebox{0.82}{\ensuremath{\boldsymbol{\times}}}}}}

\title{OmniBrain-0: A Unified Agentic Model for Embodied Intelligence}
\author{Xiaomi Embodied Intelligence Team\\
\vspace*{0.3em}
University of Macau \collabicon ~Tsinghua University  }
\contribution{See the \protect\hyperlink{contributions-heading}{\textbf{Contributions and Acknowledgments}} section for a list of contributors.}

\abstract{
General-purpose embodied intelligence requires robots to coordinate diverse skills over long horizons, track task progress, and recover from failures. Specialist policies alone do not ensure task-level reliability, while powerful multimodal agents rely on costly reasoning and external experience that compact models struggle to reproduce. We introduce \textbf{OmniBrain-0}, a lightweight unified embodied agent that coordinates manipulation and navigation through a shared high-level decision model, integrating task understanding, planning, progress tracking, and feedback-driven recovery. Our progressive training framework combines embodied pretraining, domain-specialized agent distillation, and multi-teacher on-policy distillation (MOPD). It distills powerful teachers' interaction experience into manipulation and navigation experts, then consolidates their decision capabilities into OmniBrain-0. Unification therefore occurs within the learned decision model rather than only at the tool interface. We further introduce \emph{Rubric-Guided Recursive Self-Improvement} (RSI), a data flywheel in which a Meta-Agent turns execution evidence into refined rubrics, step-level reminders, and verified trajectories for domain-expert updates before their reconsolidation into OmniBrain-0 through MOPD. We evaluate OmniBrain-0 on LIBERO-Pro, LIBERO-Plus, RoboCasa365, RoboTwin 2.0, and R2R-CE Val-Unseen, as well as multiple real-world manipulation tasks. OmniBrain-0 combines strong cross-domain task performance with substantially lower inference and execution overhead than its closed-source teacher, demonstrating efficient transfer and consolidation of embodied capabilities into a compact unified agent.
\\[1ex]
{\sffamily\bfseries Date:} \today\\
{\sffamily\bfseries Project page:} \href{https://public-bots.github.io/OmniBrain-0/}{\textsf{\bfseries https://public-bots.github.io/OmniBrain-0}}\\
\vspace{-2em}
}

\begin{document}

\maketitle
\vspace{-2em}

\begin{figure}[!htb]
\centering
\includegraphics[width=\textwidth]{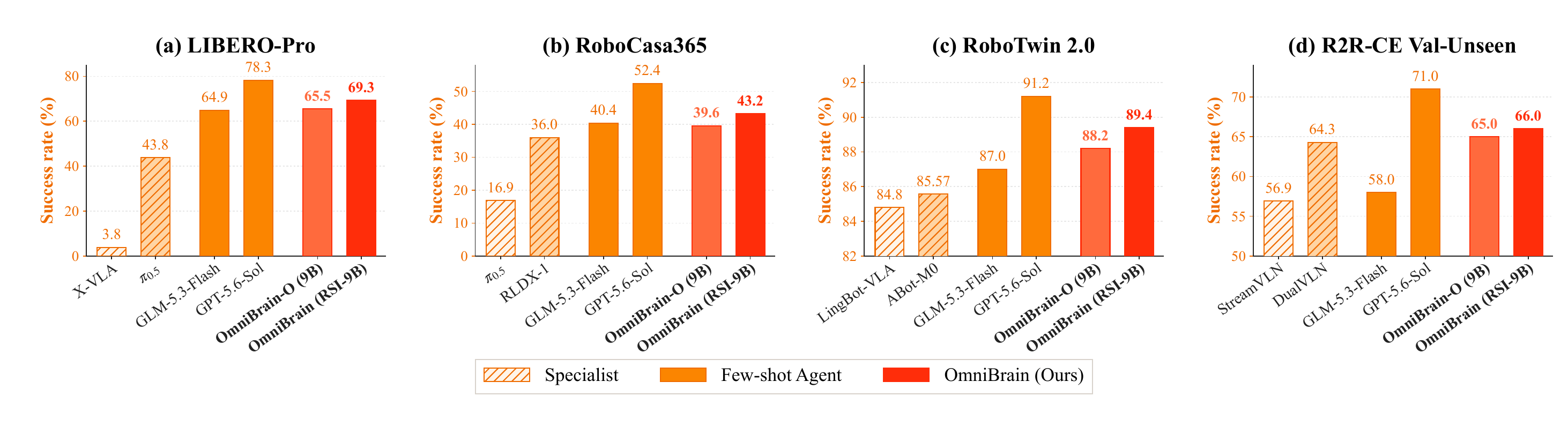}
\caption{\textbf{Cross-domain embodied-agent performance.} Success rates on LIBERO-Pro, RoboCasa365, RoboTwin~2.0, and R2R-CE Val-Unseen, comparing specialist policies, few-shot teachers, and OmniBrain-0.}
\label{fig:abstract}
\end{figure}

\clearpage
\onecolumn

\tableofcontents
\clearpage

\begingroup
\setlength{\parskip}{3.5pt plus 1pt minus 1pt}
\section{Introduction}
\label{sec:introduction}

\begin{figure}[H]
    \centering
    \includegraphics[width=1.0\textwidth]{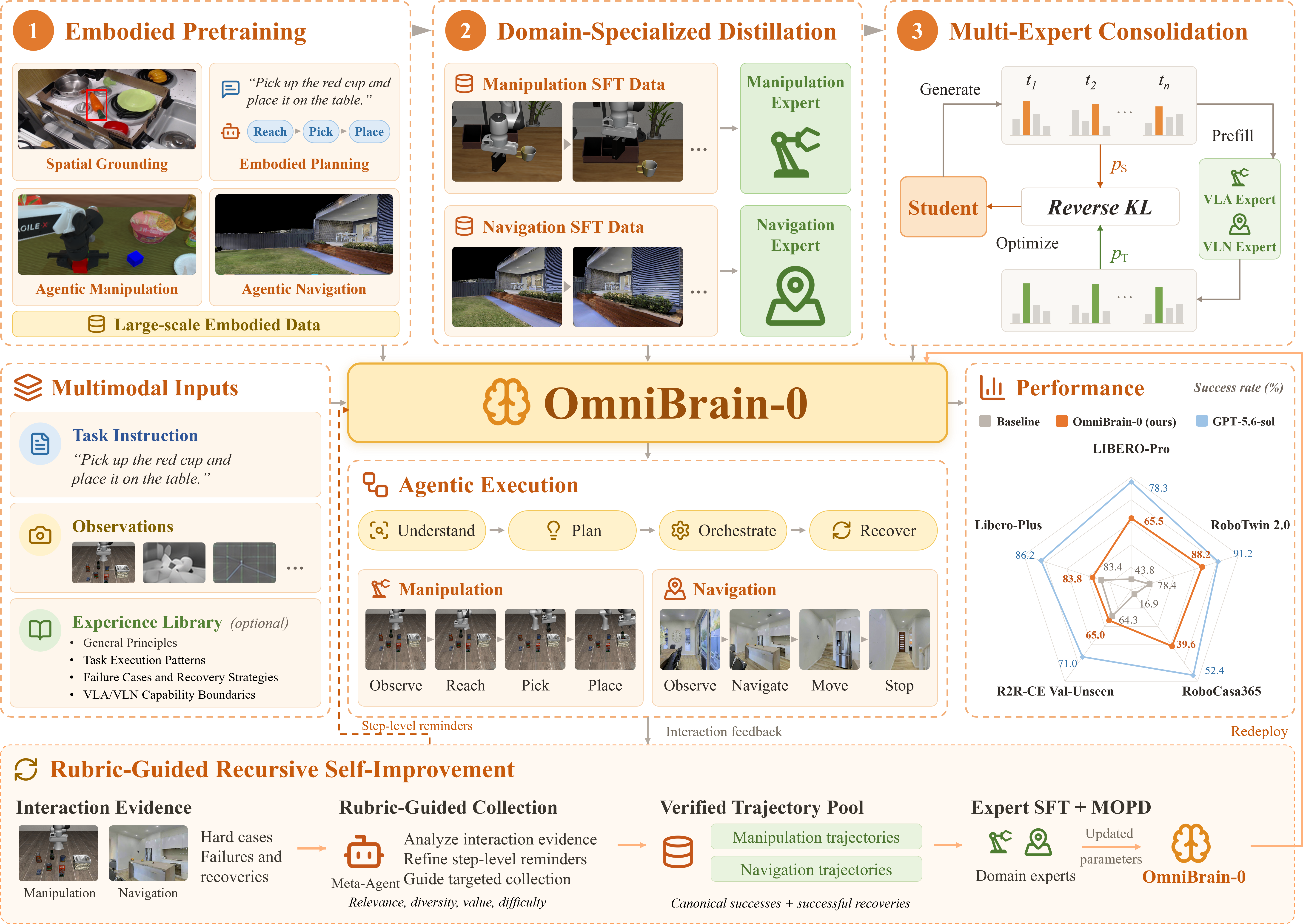}
    \caption{\textbf{Overview of \method.}
    \textit{Top:} Three-stage training: embodied pretraining, domain-specialized distillation into manipulation and navigation experts, and multi-expert consolidation (MOPD) into a single compact agent.
    \textit{Middle:} Given a task instruction, observations, and an optional experience library, \method{} understands, plans, orchestrates skills, and recovers from failures across manipulation and navigation.
    \textit{Bottom:} Rubric-Guided Recursive Self-Improvement (RSI) uses interaction evidence to refine step-level reminders and curate verified trajectories, then updates domain experts and consolidates their capabilities through MOPD for the next interaction round.
    The radar chart reports success rates (\%) on five benchmarks, comparing \method{} with the baseline policy ($\pi_{0.5}$ for LIBERO-Pro, LIBERO-Plus, RoboTwin~2.0, and RoboCasa365; InternVLA-N1 for R2R-CE Val-Unseen) and the few-shot teacher GPT-5.6-Sol.}
    \label{fig:main}
\end{figure}

General-purpose embodied intelligence requires diverse physical skills, long-horizon planning, skill composition, and closed-loop correction to fulfill task intent.
Vision-language-action (VLA) models and vision-and-language navigation (VLN) policies advance manipulation and navigation~\cite{pmlr-v270-kim25c,pmlr-v305-black25a,wei2026ground,ucagp2026}, but reliable individual skills do not guarantee task-level success.
Under complex constraints, multi-stage dependencies, or distribution shifts, local action prediction may fail to maintain goal alignment, requiring feedback-driven reassessment and replanning~\cite{zhang2026harness,xu2026worldinloop}.
Large-model-based embodied agents offer a promising approach by exposing perception, specialist policies, and motion controllers as tools.
A high-level model interprets instructions, decomposes tasks, orchestrates skills, and uses execution feedback to track progress and recover from failures, while low-level policies retain their physical interaction expertise~\cite{zhang2026harness,liang2023code,huang2022inner}.
This division of labor enables goal- and feedback-driven execution, but reliable coordination also requires the high-level model to reason about skill applicability, capability boundaries, and handoff states~\cite{huang2026roboharness}.
These shared task-level requirements motivate a common high-level decision mechanism for manipulation and navigation, spanning task understanding, skill orchestration, and feedback-driven adaptation.

However, successful task execution by powerful models does not by itself establish how their planning and recovery capabilities can be learned by a compact, unified agent.
These models combine contextual reasoning, experience, and environmental feedback~\cite{zhang2026harness,huang2026roboharness}, and compact agents must learn from this interaction experience how to decompose tasks, coordinate skills, track progress, and adapt to execution outcomes across domains.
Repeated reasoning by powerful models during long-horizon interaction incurs computational overhead and decision latency, constraining deployment efficiency~\cite{pmlr-v305-chen25a,adathinkdrive,luo2026last,lu2026xiaomi,li2025recogdrive}.
Distilling these capabilities into compact agents offers a way to improve deployment efficiency, but effective interaction data remain limited, particularly for states that require feedback-driven correction.
Useful supervision must therefore preserve how tasks progress, how skills are coordinated, and how decisions change in response to execution feedback, including during recovery from failure.

Alongside effective learning and consolidation methods, we therefore need a scalable data construction mechanism that preserves informative task progression and recovery while continually expanding coverage of execution challenges encountered during interaction~\cite{luo2026unleashing,chen2026drivereward}.
These requirements motivate a unified learning framework that combines scalable interaction data construction with capability transfer and consolidation, enabling a compact agent to coordinate planning and recovery across manipulation and navigation.

First, capability transfer aims to learn effective decision behaviors from teacher interactions rather than reproduce teachers' language outputs or reasoning processes~\cite{kang2025distillingllmagent}.
Because embodied decisions alter the environment and subsequent observations, students must learn to interpret execution feedback, track progress, and adapt when plans become invalid.
Fixed demonstrations alone may not provide sufficient supervision for appropriate decisions in states encountered during autonomous execution~\cite{huang2022inner,agarwal2024policy}.
Second, acquiring specialized capabilities and integrating them into a unified agent are distinct problems.
Manipulation and navigation differ in scene distributions, skill interfaces, and task structures, so expertise learned by specialized planners must be further integrated for joint deployment~\cite{ma2026mopd}.

Therefore, we introduce \textbf{\method}, a \textbf{lightweight, unified embodied agent for manipulation and navigation} (Figure~\ref{fig:main}).
\method~integrates specialist policies and tools through a shared skill-invocation framework.
A compact high-level planner performs task understanding, long-horizon planning, skill orchestration, state tracking, and failure recovery.
Conditioned on observations, instructions, and interaction history, it selects skills and adapts to execution feedback, preserving low-level specialization while learning task decomposition and adaptive correction.

To realize these capability transformations, we develop a progressive training framework comprising \emph{shared embodied initialization, specialized capability acquisition, and cross-domain capability integration}.
First, we combine embodied, question-answering, and general language data with limited long-horizon interaction data to establish a shared initialization for a compact model.
We then train manipulation and navigation planners from this initialization using quality-filtered interaction trajectories generated by multiple powerful models.
By learning observation--decision--feedback relationships from canonical execution and successful recovery, these planners acquire the teachers' planning and correction behaviors.
Finally, we adopt multi-teacher on-policy distillation (MOPD)~\cite{ma2026mopd}, using the domain experts to supervise student-generated decision sequences and integrate specialized capabilities into a unified student initialized from the shared pretrained model.
The resulting OmniBrain-0 thus unifies manipulation and navigation at the level of learned high-level decisions as well as the skill-invocation interface.

To support the scalable construction of effective interaction data, we introduce \emph{Rubric-Guided Recursive Self-Improvement} (RSI), a closed-loop mechanism connecting data collection, execution-guidance refinement, and model training (Section~\ref{sec:rsi}).
OmniBrain-0's autonomous interactions reveal challenges in task progression, skill coordination, and failure recovery.
A Meta-Agent analyzes execution evidence from successful and failed trajectories, derives execution rubrics, and refines step-level reminders to guide subsequent interaction and collection.
Candidate reminders are validated through interaction with model parameters fixed, while successful, failed, and inefficient executions inform further refinement.
Environment-verified canonical successes and successful recoveries are curated and structured to augment manipulation and navigation training pools, which are then used to update domain experts.
MOPD consolidates the updated experts’ capabilities into the unified model, whose subsequent interactions provide evidence for the next round of guidance refinement and data collection.
This process connects challenge discovery, execution guidance, trajectory verification, and capability learning, supporting the continued expansion of effective embodied interaction data.
We evaluate OmniBrain-0 on five manipulation and navigation benchmarks---LIBERO-Pro, LIBERO-Plus, RoboCasa365, RoboTwin~2.0, and R2R-CE Val-Unseen---with additional assessments of spatial understanding, spatial referring, and real-world manipulation.

Our main contributions are summarized as follows:
\begin{itemize}[itemsep=2pt,parsep=0pt,topsep=3pt]
    \item \textbf{A unified agentic model for manipulation and navigation.}
    We introduce OmniBrain-0, a compact high-level agent that coordinates heterogeneous perception, manipulation, navigation, and motion skills through a shared decision model for planning, orchestration, and recovery.

    \item \textbf{Scalable interaction data construction for embodied capability transfer.}
    We construct structured trajectories that preserve observation--decision--feedback dependencies and recovery, and develop a closed-loop mechanism that combines challenge-driven collection, execution-guidance refinement, and environment verification to expand training data for compact domain-specialized agents.

    \item \textbf{Progressive specialization and multi-expert consolidation.}
    Embodied pretraining, domain-specialized distillation, and MOPD unify manipulation and navigation expertise while preserving low-level specialization.
\end{itemize}

\endgroup
\section{Related Work}
\label{sec:related_work}

\subsection{Specialist policies for manipulation and navigation}
VLA models such as $\pi_0$~\citep{BlackK-RSS-25}, $\pi_{0.5}$~\citep{pmlr-v305-black25a}, and LingBot-VLA~\citep{wu2026pragmatic,wu2026foundation} learn multitask control from large-scale robot data and vision-language pretraining.
These policies provide a foundation for reusable manipulation skills~\citep{ucagp2026}, while task-level coordination must also account for sub-goal dependencies, execution progress, and the conditions required by subsequent skills.
Yet even $\pi_{0.5}$ and LingBot-VLA~2.0 degrade substantially on long-horizon tasks under object and initial-position shifts~\citep{wu2026foundation}.
For navigation, NavGPT-2~\citep{zhou2024navgpt} couples language reasoning with a navigation policy, and DualVLN (InternVLA-N1)~\citep{wei2026ground} connects high-level planning to low-level execution in a dual-system design.
Manipulation and navigation nevertheless involve different observations, action interfaces, and temporal structures, motivating a shared decision layer that can coordinate domain-specific execution.
Rather than replacing these policies, \method{} invokes them as skills and learns a shared high-level decision model across both domains.

\subsection{Tool-using embodied agents}
SayCan~\citep{ahn2022can} and Code as Policies~\citep{liang2023code} ground task decomposition in executable skills, and Inner Monologue~\citep{huang2022inner} feeds environmental feedback back into planning.
This division between task reasoning and skill execution provides a basis for composing heterogeneous capabilities, with feedback enabling the planner to reassess whether intermediate outcomes satisfy the current goal~\citep{mtrdrive}.
Agent harnesses, including Harness VLA~\citep{zhang2026harness}, RoboHarness~\citep{huang2026roboharness}, and Pigey~\citep{galanti2026addressing}, add state tracking, outcome verification, and recovery through agent loops.
However, their orchestration relies on repeated reasoning by large models at deployment time.
\method{} instead internalizes planning, progress tracking, and recovery into a compact decision model learned from interaction experience.
Building on this execution paradigm, we retain the tool interfaces and feedback loop while using curated interaction trajectories to supervise the high-level decisions made within it.

\subsection{Agent distillation and expert consolidation}
AgentTuning~\citep{zeng2024agenttuning} and STEVE-2~\citep{zhao2024we} transfer agent capabilities via trajectory tuning and hierarchical distillation, but fixed teacher trajectories may not cover states encountered by students.
For embodied capability transfer, useful supervision must preserve the relationship between observations, tool choices, and execution outcomes, including corrective decisions that lead to successful recovery.
On-policy distillation mitigates this mismatch by supervising student-generated sequences~\citep{agarwal2024policy,luo2026reflectvla}; TurnOPD~\citep{zhou2026turnopd} adds turn-aware supervision for agents, and MOPD~\citep{ma2026mopd} merges domain-specialized teachers.
Such consolidation complements trajectory-based specialization: the latter acquires domain-specific decision behaviors, while the former integrates complementary expert distributions within shared model parameters.
In embodied settings, each decision alters subsequent observations and skill applicability; \method{} applies multi-teacher distillation to consolidate manipulation and navigation expertise, including planning and recovery, into a single model.

\subsection{Recursive self-improvement of embodied agents}
\label{sec:related_rsi}
Recursive self-improvement (RSI) uses successive interactions to refine an agent's guidance or learned policy.
Reflexion~\citep{shinn2023reflexion} stores verbal reflections in episodic memory, while ACE~\citep{zhang2026ace} evolves contextual playbooks through reflection and curation.
ReST meets ReAct~\citep{aksitov2023rest} iteratively trains agents on feedback-filtered trajectories.
In robotics, EmbodiRSI~\citep{lang2026embodirsi} targets corrective data collection at policy weaknesses, and AdaRSI~\citep{wang2026adarsi} combines harness refinement with selective executor fine-tuning.
RE-0~\citep{zhang2026re0} verifies local teacher corrections before distilling them into embodied code-generating agents.
\method{} connects rubric-guided reminder refinement with verified trajectory curation, domain-expert SFT, and MOPD for a decision model shared across manipulation and navigation.
Reminders are validated without changing model parameters; subsequent training updates the agent, whose new interactions guide the next improvement round.
This cycle retains successful recoveries while using failed episodes for diagnosis.

\section{Unified Embodied Agent Architecture}
\label{sec:method}

\begin{figure}[!htbp]
    \centering
    \includegraphics[width=1.0\textwidth]{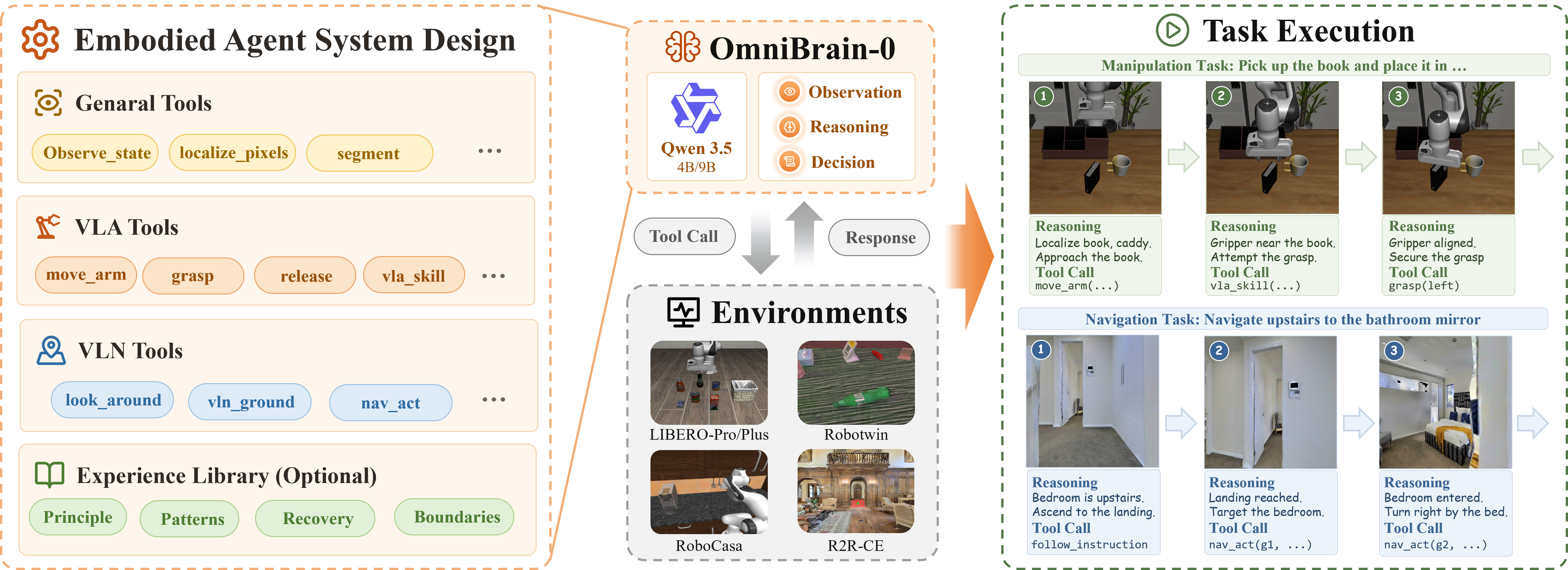}
    
    \caption{\textbf{Agentic execution framework and task execution examples.}
    \textit{Left:} Unified agentic system design comprising general perception tools, VLA manipulation tools, VLN navigation tools, and an optional experience library.
    \textit{Center:} \method{}~performs observation-driven reasoning and tool orchestration through closed-loop interaction with task environments, using execution feedback to guide subsequent decisions.
    \textit{Right:} Representative manipulation and navigation examples illustrating step-by-step reasoning, tool invocation, and task progression.}

    \label{fig:method}
\end{figure}

\subsection{Problem Formulation and Overview}
\label{sec:problem_formulation}

We consider language-conditioned manipulation and navigation tasks, with tool invocation serving as a unified high-level action representation.
Given a task instruction $g$, OmniBrain-0 selects tools based on current observations and interaction history, continually updating its decisions through environmental feedback.
At decision step $t$, the observation and interaction context are defined as
\begin{equation}
    o_t=(I_t,q_t,p_t),
    \qquad
    \mathcal H_t=(h_t,\mathcal M),
    \label{eq:observation_context}
\end{equation}
where $I_t$ denotes visual observations, $q_t$ the robot's proprioceptive state, and $p_t$ task-status information exposed by the environment.
The within-episode interaction history is denoted by $h_t$, while $\mathcal M$ represents an optional experience library.
Decisions are conditioned only on information available at the current or previous steps, with the experience library providing priors for task organization and skill use.

Let $\mathcal P_e$ denote the available tool set, the high-level policy $\pi_\theta$ generates a tool identifier $k_t\in\mathcal P_e$ and invocation parameters $\xi_t$:
\begin{equation}
    \begin{aligned}
        x_t &= (g,o_t,\mathcal H_t),\\
        u_t &= (k_t,\xi_t)\sim\pi_\theta(\cdot\mid x_t),
        \qquad k_t\in\mathcal P_e.
    \end{aligned}
    \label{eq:skill_invocation}
\end{equation}
where $u_t$ the executed skill invocation. The selected tool performs the corresponding perception or control operation and returns a new observation $o_{t+1}$ and execution feedback $f_{t+1}$, which condition subsequent decisions.
The objective is to fulfill the instruction within a given execution budget, with completion determined by the environment's task criteria.
This formulation provides a unified high-level tool interface for manipulation and navigation, while preserving their distinct low-level action spaces and execution durations.

\subsection{Skill Interfaces and Context Organization}
\label{sec:architecture}
The agentic execution framework of \method{} follows the system design of Harness VLA~\citep{zhang2026harness}, with its tools and experience library summarized in Figure~\ref{fig:method} (left).
OmniBrain-0 uses the general-purpose multimodal model Qwen3.5-9B~\cite{qwen3.5} as a shared agent, decoupling task-level decision-making from specialized tool execution. The agent performs instruction understanding, long-horizon task decomposition, tool orchestration, and failure recovery, while execution modules provide perception, manipulation, navigation, and motion-control capabilities. A unified invocation-and-feedback interface connects the agent and execution modules in a closed loop, enabling the same decision model to coordinate heterogeneous tools with different inputs and control mechanisms.

\paragraph{Unified tool abstraction}
We encapsulate heterogeneous execution capabilities as parameterized semantic tools. Perception tools provide scene understanding and target localization; VLA tools handle manipulation and contact interactions; navigation tools support goal reaching through VLN policies or geometric navigators; and motion-control tools provide explicit pose and actuator adjustments. Each tool retains its native conditioning and low-level implementation while exposing a unified invocation-and-feedback interface. Thus, the agent shares a common high-level decision mechanism and interaction protocol rather than a single low-level control policy.

\paragraph{Context-driven tool orchestration}
The agent jointly processes task instructions, current observations, and historical feedback to select subsequent tools and determine their invocation parameters. Interaction history supports task progress tracking, while an optional experience library provides priors on task organization, tool composition, and recovery strategies. Experience guides task-level organization, whereas online perception grounds tool invocation in the current execution state.

\subsection{Closed-Loop Planning and Failure Recovery}
\label{sec:closed_loop_execution}
At each turn, the agent selects one tool and makes the next decision only after receiving its execution feedback. High-level tool invocations are executed sequentially and connected through execution outcomes. The agent continuously updates its decisions based on observations and feedback, adapting target localization, invocation parameters, or tool selection when execution deviates from expectations. Detailed tool definitions and implementation specifications are provided in the Appendix~\ref{appendix:agent_system}.

\section{Embodied Capability Learning and Consolidation}
\label{sec:training}
\label{sec:capability_transfer}

\begin{figure}[!htbp]
    \centering
    \includegraphics[width=\textwidth]{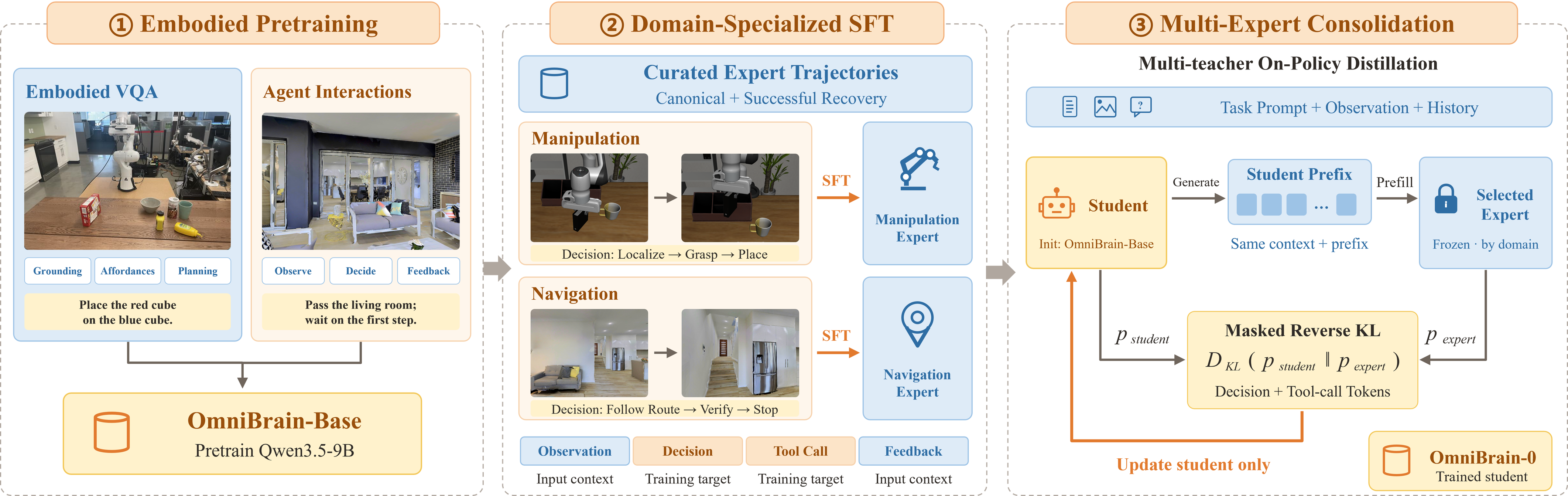}
    \caption{\textbf{Three-stage training framework of \method.}
    \textit{Left:} Embodied VQA and agent interaction data establish
    the shared OmniBrain-Base initialization.
    \textit{Middle:} Curated expert trajectories, including canonical
    successes and successful recoveries, supervise domain-specialized
    manipulation and navigation agents.
    \textit{Right:} MOPD consolidates the frozen domain experts into
    a unified student through masked reverse-KL supervision on
    student-generated decision prefixes, updating only the student.}
    \label{fig:training_framework}
\end{figure}

Figure~\ref{fig:training_framework} summarizes our three-stage capability transfer framework: embodied pretraining establishes a shared initialization (Section~\ref{sec:embodied_pretraining}), domain-specialized distillation transfers expert interaction experience into specialized decision-making capabilities (Section~\ref{sec:specialized_distillation}), and multi-teacher on-policy distillation consolidates these capabilities into a unified shared model (Section~\ref{sec:mopd}). This process separates capability acquisition from cross-domain integration, allowing the shared agent to inherit complementary planning, tool orchestration, and recovery capabilities while preserving the specialization of low-level execution tools.

\label{sec:progressive_training}

\subsection{Interaction Data Construction (Data Pipeline)}
\label{sec:data_pipeline}

\begin{figure}[!htbp]
    \centering
    \includegraphics[width=1.0\textwidth]{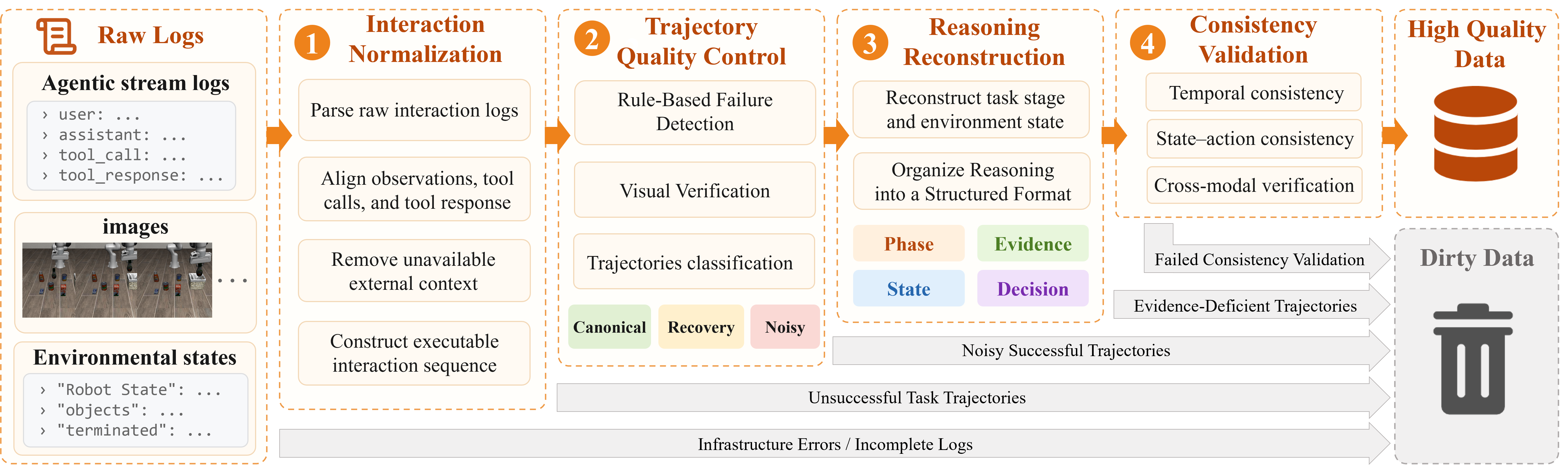}
    \caption{\textbf{Interaction data construction.} Raw interaction logs, images, and environment states pass through interaction normalization, trajectory quality control, reasoning reconstruction, and consistency validation. Canonical successes and successful recoveries form the training data; incomplete, unsuccessful, noisy, evidence-deficient, or inconsistent trajectories are discarded.}
    \label{fig:data_pipeline}
\end{figure}

\label{appendix:data_pipline}
\paragraph{Expert Interaction Trajectories}
\label{sec:teacher_trajectories}
We collect expert demonstration data by leveraging multiple powerful models to execute manipulation and navigation tasks in closed-loop environments.
Experts invoke tools within an execution framework that a Meta-Agent iteratively refines, continually adjusting decisions based on task objectives, online observations, and execution feedback.
Each trajectory is organized as a temporally ordered, multi-turn multimodal sequence:
\begin{equation}
    \tau=\left(
        g,x_0,y_0,u_0,f_1,x_1,\ldots,
        y_{T-1},u_{T-1},f_T,x_T
    \right),
    \label{eq:teacher_trajectory}
\end{equation}
where $y_t$ denotes the expert's decision text, $u_t$ the executed skill invocation, and $f_{t+1}$ the corresponding execution feedback; $x_t$ follows the definition in Section~\ref{sec:problem_formulation}.
We preserve the temporal relationships among observations, decisions, and feedback, ensuring that each decision is grounded in information available at that step.

Figure~\ref{fig:data_pipeline} details the conversion of expert interaction records into training trajectories through four steps: \textit{interaction normalization}, \textit{trajectory quality control}, \textit{reasoning reconstruction}, and \textit{consistency validation}. The pipeline aligns raw logs, images, and environment states into supervision compatible with the student's observations and tool interfaces. It preserves the observation--action--feedback structure and retains both canonical successes and successful recoveries, while removing task-irrelevant information and unreliable trajectories. Dataset composition and splits are provided in the experimental setup and appendix.

\paragraph{Interaction Normalization}
We parse raw interaction logs and reconstruct each episode as a temporally ordered sequence of \textit{reasoning--tool call--tool feedback} units, retaining synchronized images and robot states at each decision step. Tool calls and feedback are normalized into a unified format, with visual observations aligned to the corresponding interaction steps. Context unavailable to the student is removed, while infrastructure errors and incomplete logs are filtered out to obtain executable interaction sequences.

\Needspace{4\baselineskip}
\paragraph{Trajectory Quality Control}
We use rule-based failure detection and visual verification to assess task outcomes and execution quality, discarding unsuccessful task trajectories. Successful trajectories do not necessarily provide equal training value. Canonical successful trajectories provide reliable demonstrations of standard task execution, while failure-recovery trajectories offer valuable state-dependent supervision by capturing how the agent diagnoses and corrects execution failures. In contrast, blind repetitions without new observations or state information may reinforce inefficient action patterns. We therefore categorize successful trajectories into three types:

\begin{itemize}
\item \textbf{Canonical trajectories}: The task is completed without tool failures, with perception, manipulation, and termination proceeding as expected.
\item \textbf{Recovery trajectories}: One or more grasping, contact, or other execution failures occur, but the agent subsequently recovers through re-localization, gripper adjustment, re-grasping, or other corrective actions and eventually receives an environment success signal.
\item \textbf{Noisy trajectories}: The agent repeatedly executes the same action without acquiring new observations or state information, reflecting blind retries rather than feedback-driven decision making.
\end{itemize}

Both canonical and recovery trajectories are retained, while noisy trajectories are discarded. Canonical trajectories provide standard execution demonstrations, while recovery trajectories capture the failure--diagnosis--correction--continuation process.

\paragraph{Reasoning Reconstruction}
The raw reasoning text generated by the expert model may introduce unnecessary linguistic variation. We therefore rewrite each reasoning step into a structured representation using only information available at the corresponding decision step, following two principles: \textbf{state grounding}, which derives the state from tool outputs, robot states, and images, and \textbf{evidence traceability}, which excludes future information. Specifically, each reasoning step is represented by four fields:

\begin{itemize}
\item \textbf{Phase}: the current stage of task execution, such as localization, grasping, transport, placement, release, success verification, or recovery.
\item \textbf{State}: the current interaction state, including the held object, gripper and grasp status, target and destination locations, and task completion status.
\item \textbf{Evidence}: relevant visual observations and tool feedback supporting the current state assessment.
\item \textbf{Decision}: the next tool to be invoked and its action description determined by the actual tool parameters.
\end{itemize}

We generate these fields by replaying each trajectory chronologically: tool feedback first updates the interaction state, while visual auditing verifies information not reliably available from tool outputs, such as object identity, grasp status, and placement. The \textit{Phase}, relevant \textit{Evidence}, and \textit{Decision} are then determined from the resulting state and the actual next tool call. Failed actions are marked as recovery operations, while environment-confirmed completion produces the terminal state. For long-horizon tasks, the \textit{State} additionally tracks the current sub-goal and its completion progress. The resulting structured reasoning is independent of the teacher's wording and remains one-to-one aligned with each tool call.

\paragraph{Consistency Validation}
We finally check temporal consistency, state--action consistency, and agreement between visual observations and textual evidence. These checks verify consistency among the state, evidence, phase, and tool call, including target/destination localization, action--decision alignment, and environment-verified termination. Evidence-deficient trajectories and those failing consistency validation are discarded, while canonical successes and successful recovery trajectories are retained as high-quality training data.

\subsection{Embodied Pretraining}
\label{sec:embodied_pretraining}
As shown in Figure~\ref{fig:training_framework} (left), we pretrain on a mixture of embodied question-answering data and long-horizon agent interaction data to establish a shared basis for embodied capabilities in manipulation and navigation.
Embodied question answering provides supervision for scene understanding and task-related knowledge, while long-horizon interactions guide the model to organize decisions and use feedback in multi-turn tasks through temporal relationships among instructions, observations, tool invocations, and execution feedback.
These complementary sources connect static embodied understanding with dynamic interactive decision-making, providing shared knowledge and interaction priors for subsequent specialization.
The resulting model, denoted as \textbf{OmniBrain-Base} with parameters $\theta_{\mathrm{base}}$, initializes both the domain teachers and the unified student.

\subsection{Domain-Specialized Agent Distillation}
\label{sec:specialized_distillation}
As shown in Figure~\ref{fig:training_framework} (middle), we initialize domain-specialized teachers for manipulation and navigation from OmniBrain-Base and train them through supervised hard distillation on expert interaction trajectories.
Each teacher learns goal understanding, skill orchestration, and recovery decisions appropriate to its domain's task structure.
These teachers are high-level agent models: they learn to organize and use skills rather than generate low-level actions.

Let $\phi_k$ denote the parameters of teacher $k$ and $\mathcal D_k$ its training dataset.
Given a context $x$ comprising task instructions, observations, interaction history, and skill definitions, we serialize the teacher's decision text and executed invocation into a target sequence $z$ and minimize the autoregressive negative log-likelihood:
\begin{equation}
    \begin{aligned}
        \mathcal L_{\mathrm{expert}}^{(k)}
        &=-\mathbb E_{(x,z)\sim\mathcal D_k}
        \left[
            \frac{1}{|z|}
            \sum_{j=1}^{|z|}
            \log\pi_{\phi_k}(z_j\mid x,z_{<j})
        \right],\\
        \phi_k^{(0)}&=\theta_{\mathrm{base}}.
    \end{aligned}
    \label{eq:expert_distillation}
\end{equation}
Supervision is applied only to the agent's decision text and tool invocations, while environmental observations and tool feedback serve as conditioning information.
By jointly learning from standard execution and recovery trajectories, specialization transforms expert interaction experience into observation-conditioned, task-level decision-making capabilities, yielding domain teachers for subsequent cross-domain integration.

\subsection{Multi-Expert Consolidation with MOPD}
\label{sec:mopd}
As shown in Figure~\ref{fig:training_framework} (right), we adopt multi-teacher on-policy distillation (MOPD)~\cite{ma2026mopd} to integrate domain teachers' distributional supervision over student-generated decision prefixes.
The unified student is initialized from OmniBrain-Base, and the domain-specialized agents obtained in Section~\ref{sec:specialized_distillation} form a fixed teacher set:
\begin{equation}
    \theta^{(0)}=\theta_{\mathrm{base}},
    \qquad
    \mathcal T=\{\pi_{\phi_k}\}_{k=1}^{K},
    \label{eq:consolidation_initialization}
\end{equation}
where $K$ is the number of teachers.
The student and teachers share the same model architecture and vocabulary.
Each teacher provides domain-specific supervision, jointly optimizing the unified student.

\noindent\textbf{Teacher Supervision on Student Prefixes.}
We sample domain $k$ with weight $\rho_k$, where $\rho_k\geq0$ and $\sum_k\rho_k=1$.
For a domain context $x\sim\mu_k$, a student decision sequence is sampled as $\hat z\sim\pi_{\bar\theta}(\cdot\mid x)$, where $\bar\theta$ denotes the student parameters used for sampling. The teacher provides token-level distributional supervision conditioned on the same context and the student-generated prefix:
\begin{equation}
    \begin{aligned}
        p_{\theta,j}(v)
        &=\pi_\theta(v\mid x,\hat z_{<j}),\\
        q_{k,j}(v)
        &=\pi_{\phi_k}(v\mid x,\hat z_{<j}),
        \qquad v\in\mathcal V,
    \end{aligned}
    \label{eq:expert_student_distributions}
\end{equation}
where $\mathcal V$ denotes the shared top-$K$ token support with renormalized student and teacher distributions.
At this stage, distribution alignment is conditioned on student-generated prefixes, allowing teacher supervision to adapt to the student's generation behavior.

\medskip
\noindent\textbf{Joint multi-domain distillation objective.}
We minimize the token-level reverse Kullback--Leibler (KL) divergence from the student to the corresponding expert:
\begin{equation}
    \begin{aligned}
        \mathcal L_{\mathrm{MOPD}}(\theta)
        & = \sum_{k=1}^{K}\rho_k
        \mathbb E_{\substack{
            x\sim\mu_k\\
            \hat z\sim\pi_{\bar\theta}(\cdot\mid x)
        }}
        \left[
            \frac{1}{M}
            \sum_{j=1}^{|\hat z|}
            m_j\,
            D_{\mathrm{KL}}\!\left(
                p_{\theta,j}\,\middle\|\,q_{k,j}
            \right)
        \right],\\
        M & = \sum_{j=1}^{|\hat z|}m_j>0,
    \end{aligned}
    \label{eq:mopd_objective}
\end{equation}
where $m_j\in\{0,1\}$ marks only valid agent decisions and tool-invocation outputs.
The divergence is defined as
\begin{equation}
    D_{\mathrm{KL}}\!\left(
        p_{\theta,j}\,\middle\|\,q_{k,j}
    \right)
    =
    \sum_{v\in\mathcal V}
    p_{\theta,j}(v)
    \log\frac{p_{\theta,j}(v)}{q_{k,j}(v)}.
    \label{eq:reverse_kl}
\end{equation}
This objective weights distributional discrepancies by student prediction probabilities, using the decision preferences of different teachers to constrain the student's current output distribution.
Teacher parameters and sampled prefixes remain fixed during optimization, and only the unified student parameters $\theta$ are updated.
By decoupling specialization from cross-domain integration, we consolidate manipulation and navigation expertise into the lightweight \textbf{OmniBrain-0}, which orchestrates cross-domain skills with a single set of high-level agent parameters.

\section{Rubric-Guided Recursive Self-Improvement}
\label{sec:rsi}

We introduce \emph{Rubric-Guided Recursive Self-Improvement} (RSI) to support the scalable construction of effective embodied interaction data.
A Meta-Agent derives step-level guidance from interaction evidence, while verified trajectories augment domain-expert SFT.
MOPD consolidates the updated experts into OmniBrain-0, whose subsequent interactions support the next round of improvement.

\subsection{Rubric-Guided Reminder Policy}
\label{sec:reminder_policy_iteration}

\begin{figure}[!htbp]
    \centering
    \includegraphics[width=\linewidth]{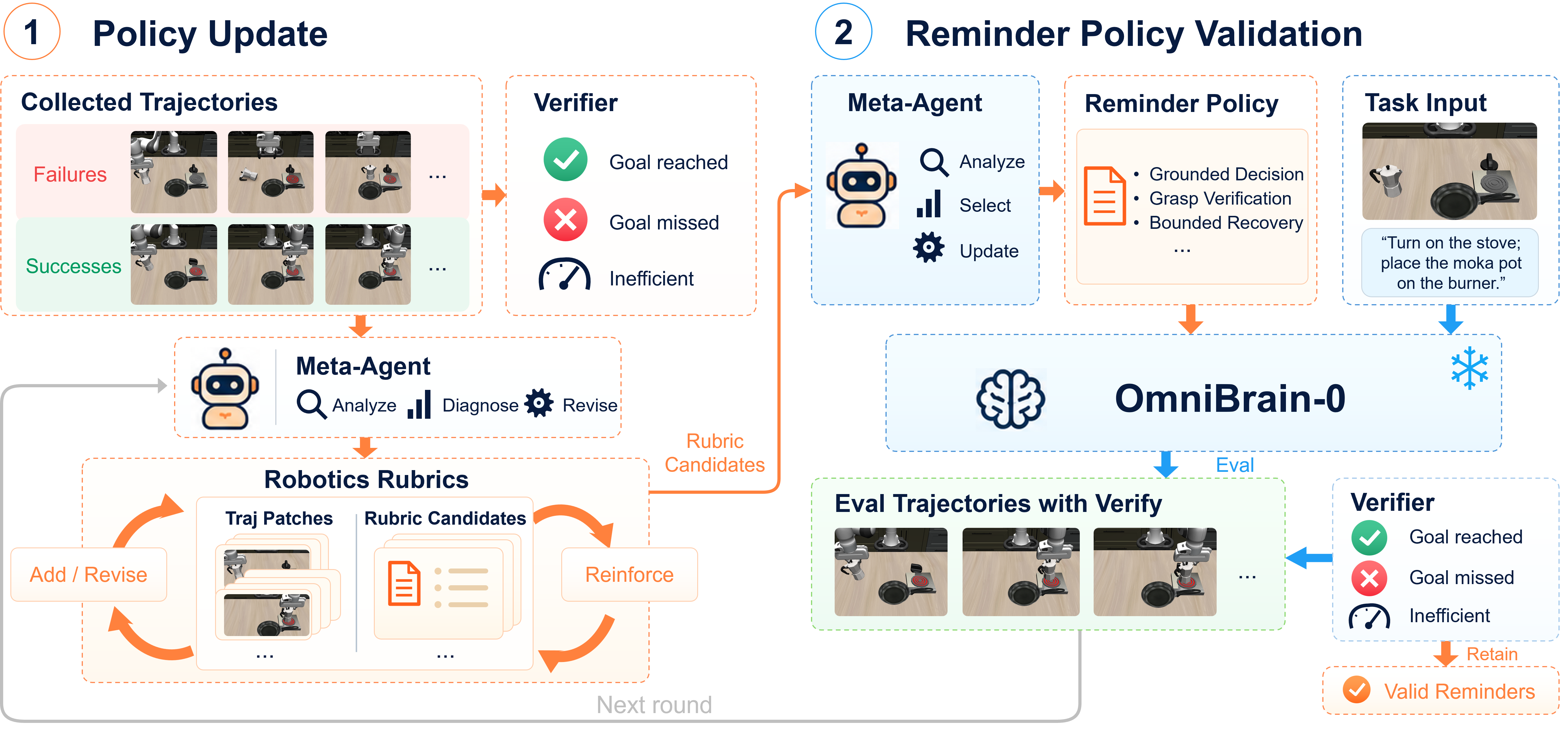}
    \caption{\textbf{Rubric-based policy update and reminder-policy validation.} \textit{Left:} A verifier assesses collected successful and failed trajectories. The Meta-Agent analyzes execution evidence and adds, revises, or reinforces robotics rubrics supported by trajectory patches. \textit{Right:} Rubric candidates inform the reminder policy, which conditions OmniBrain-0 together with the task input while model parameters remain fixed. Verified evaluation trajectories provide feedback for the next round, and effective reminders are retained as valid reminders.}
    \label{fig:rsi_reminder}
\end{figure}

\paragraph{Rubric-based policy update}
Let $\mathcal D_k$ denote the self-collected trajectories at refinement round $k$, following the interaction representation in Equation~\ref{eq:teacher_trajectory}.
The initial collection includes both successful and failed manipulation and navigation episodes.
A verifier assesses goal attainment and execution efficiency, providing outcome evidence for the Meta-Agent to analyze, diagnose, and revise execution guidance.
Successful trajectories reveal effective task progression, while failed and inefficient trajectories expose unresolved execution challenges.
The Meta-Agent relates these outcomes to the corresponding observations, decisions, and feedback, organizing relevant trajectory patches and rubric candidates into a shared set of robotics rubrics $\mathcal U_k$.
New evidence supports adding or revising rubrics, while recurring supporting evidence reinforces applicable criteria.
These rubrics capture principles for task prerequisites, skill ordering, progress verification, and recovery.
Figure~\ref{fig:rsi_reminder} summarizes rubric-based policy update and reminder-policy validation.

Given the collected evidence and current rubrics, the update is expressed as
\begin{equation}
    \mathcal U_{k+1}=\mathcal A(\mathcal D_k,\mathcal U_k),
    \qquad
    \mathcal R_{k+1}=\mathcal G(\mathcal U_{k+1}),
    \label{eq:rubric_policy_refinement}
\end{equation}
where $\mathcal A$ denotes rubric analysis and revision, and $\mathcal G$ denotes reminder-policy construction, including candidate selection and validation.
The rubrics organize execution criteria; the reminder policy translates selected criteria into concise guidance for the agent.
The update changes the guidance supplied by the harness and retains the tool interfaces in Section~\ref{sec:architecture}.

\paragraph{Step-level reminder guidance}
The Meta-Agent analyzes rubric candidates, selects relevant guidance, and updates the reminder policy.
At every high-level decision step, the harness evaluates the current interaction context to determine whether guidance is needed.
A reminder is inserted before the next tool invocation only when an issue is detected; otherwise, the agent proceeds without a reminder.
Using the context $x_t=(g,o_t,\mathcal H_t)$ from Section~\ref{sec:problem_formulation}, the reminder and conditional decision are
\begin{equation}
    r_t=\mathcal R_{k+1}(\mathcal U_{k+1},x_t),
    \qquad
    u_t\sim\pi_\theta(\cdot\mid x_t,r_t),
    \label{eq:reminder_guidance}
\end{equation}
where $r_t$ is empty when no issue is detected and otherwise provides task- and state-dependent guidance on the prerequisites, action order, or execution evidence relevant to the current decision.
Task input and reminder guidance jointly condition tool selection and invocation parameters.
Subsequent observations and feedback update the interaction context, allowing guidance to remain aligned with task progress throughout the episode.

\paragraph{Reminder-policy validation}
OmniBrain-0 executes evaluation trajectories under the candidate guidance with its parameters fixed.
The verifier assesses goal attainment and execution efficiency, and the Meta-Agent uses the resulting evidence to identify valid reminders and revise ineffective guidance.
Verified evaluation trajectories, including successful and failed executions, are returned for the next round of rubric-based policy update.
Effective reminders are retained across rounds, enabling execution guidance to accumulate from observed interaction outcomes.
The verified successful trajectories also provide candidates for the training-data process in Section~\ref{sec:rsi_data_flywheel}.
This stage refines the reminder policy through interaction feedback; model-parameter updates are performed through expert training and MOPD below.

\subsection{RSI Data Flywheel}
\label{sec:rsi_data_flywheel}

\begin{figure}[!htbp]
    \centering
    \includegraphics[width=\linewidth]{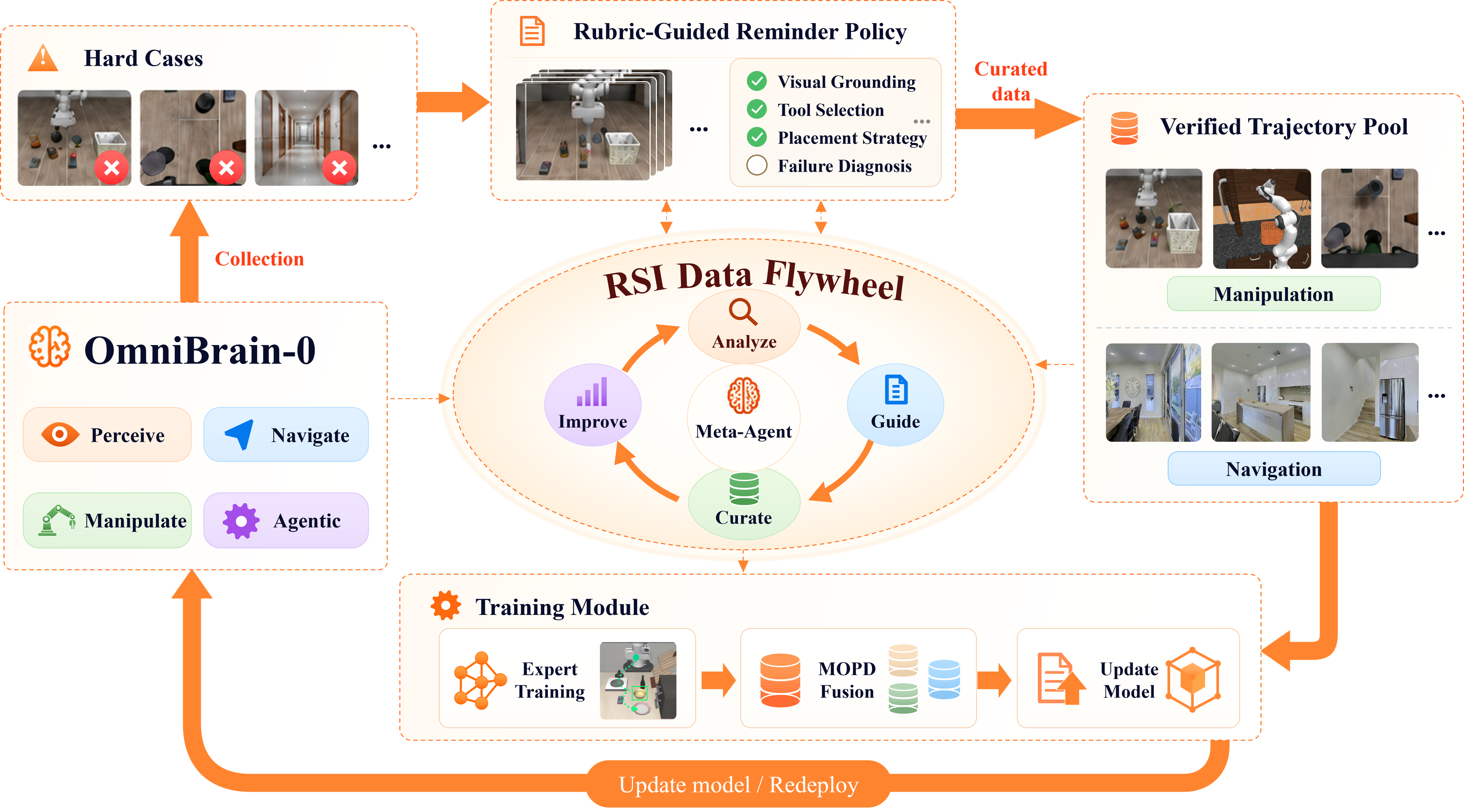}
    \caption{\textbf{RSI data flywheel.} OmniBrain-0 collects challenging cases that inform the rubric-guided reminder policy and trajectory curation. The Meta-Agent coordinates analysis, guidance, curation, and improvement, with task relevance, diversity, learning value, and difficulty considered during selection. Verified trajectories form manipulation and navigation pools for expert training. MOPD consolidates the updated experts into the unified model, which is redeployed for the next collection round.}
    \label{fig:rsi_flywheel}
\end{figure}

The RSI data flywheel connects reminder-guided interaction, verified trajectory collection, and the training framework in Section~\ref{sec:training}.
As illustrated in Figure~\ref{fig:rsi_flywheel}, the Meta-Agent coordinates analysis, guidance, curation, and improvement.
Collected experience refines the reminder policy and enriches expert training, while the updated unified model generates the next round of interaction evidence.

\paragraph{Rubric-guided trajectory collection}
OmniBrain-0 interactions reveal challenging cases that expose limitations in task progression, skill coordination, or recovery.
These cases guide subsequent collection and reminder-policy refinement.
The rubric-guided reminder policy from Section~\ref{sec:reminder_policy_iteration} supports further interaction on the identified challenges, while trajectory selection considers task relevance, diversity, learning value, and difficulty.
Together, guidance and curation focus collection on informative experience aligned with the current model's capability needs.

\paragraph{Verified trajectory pool}
Collected trajectories are verified and curated into manipulation and navigation pools.
Failed and inefficient executions remain available for diagnosis, while expert training retains environment-verified canonical successes and successful recoveries under Section~\ref{sec:data_pipeline}.
The existing data pipeline normalizes interaction records and constructs consistent, state-grounded supervision, preserving the temporal alignment of observations, decisions, tool invocations, and feedback.
This connects challenging cases to useful training experience through verified execution and data curation.

Let $\mathcal T_k^{(d)}$ denote trajectories collected for domain $d$ at round $k$, and let $\mathcal Q$ denote verification, curation, and training-data processing.
The selected trajectories augment the corresponding expert corpus:
\begin{equation}
    \mathcal D_{\mathrm{SFT},k+1}^{(d)}
    =\mathcal D_{\mathrm{SFT},k}^{(d)}
    \cup\mathcal Q\!\left(\mathcal T_k^{(d)}\right),
    \label{eq:rsi_data_feedback}
\end{equation}
where $d$ indexes manipulation or navigation.
Following Section~\ref{sec:specialized_distillation}, decision text and tool invocations are prediction targets, while observations and environment feedback provide conditioning information.
The curated trajectories therefore return to the same expert-training stage as the original interaction data.

\paragraph{Expert training and MOPD consolidation}
Domain-specialized experts are further trained on the augmented corpora using Equation~\ref{eq:expert_distillation}.
The resulting experts then provide domain-specific distributional supervision for MOPD under Equation~\ref{eq:mopd_objective}.
The MOPD fusion stage in Figure~\ref{fig:rsi_flywheel} follows this distillation procedure, transferring the refined experts' decision capabilities into the unified OmniBrain-0.
The updated model is redeployed for further collection, so subsequent challenges reflect its current capabilities.

\paragraph{Iterative improvement}
New interactions supply the next evidence batch $\mathcal D_{k+1}$ for rubric revision and reminder-policy validation.
Verified trajectories extend the expert corpora through Equation~\ref{eq:rsi_data_feedback}, while expert SFT and MOPD convert the selected experience into updated model parameters.
The flywheel thus connects \textit{analyze}, \textit{guide}, \textit{curate}, and \textit{improve}: interaction evidence informs guidance, guidance supports collection, curation yields verified supervision, and training returns an updated model to interaction.
Reminder-policy refinement and model learning remain connected through this shared flow of execution evidence and training data.

\FloatBarrier

\section{Experiments and Analysis}
\label{sec:experiments}
This section describes the evaluation setup, analyzes task performance and inference efficiency, and presents ablations and qualitative examples of task execution and failure recovery.

\subsection{Experimental Setup}

\subsubsection{Evaluation Benchmarks}
We evaluate manipulation on LIBERO-Pro~\cite{zhou2025liberopro}, LIBERO-Plus~\cite{fei2025liberoplus}, RoboCasa365~\cite{nasiriany2026robocasa365}, and RoboTwin~2.0~\cite{chen2025robotwin}; navigation on R2R-CE~\cite{krantz2020beyond}; and spatial understanding and referring on EmbSpatial-Bench~\cite{du2024embspatial} and RefSpatial-Bench~\cite{zhou2026roborefer}. Physical tests cover block stacking, apple placement, and drawer opening. We report manipulation success rates, R2R-CE SR/SPL/OSR/NE on a fixed 100-episode Val-Unseen subset, and spatial accuracy. Appendix~\ref{appendix: eval} details task splits, episode counts, perturbations, and scoring.

\subsubsection{Implementation Details}
OmniBrain-0 uses Qwen3.5-9B to coordinate perception, manipulation, and navigation tools (Section~\ref{sec:architecture}). Training combines embodied VQA and expert interaction trajectories through the pipeline and training stages in Sections~\ref{sec:data_pipeline} and~\ref{sec:training}, with additional verified trajectories from RSI. Comparisons match task settings, tools, low-level skills, observation interfaces, and execution budgets; simulated evaluations also match episodes, seeds, and simulator versions. Appendix~\ref{appendix:data_and_train} details data, collection and execution settings, Meta-Agent and judge-model assignments, hardware, and training hyperparameters.

\subsection{Overall Performance}
\label{sec:overall_performance}

Unless otherwise specified, the quantitative analysis below refers to \method{} (9B). The additional 4B and RSI rows report results for the smaller model and the model iterated through the RSI Data Flywheel, respectively. All main-table results for these three variants are obtained without an external Experience Library at evaluation time. In Tables~\ref{tab:libero_pro_perturbation}--\ref{tab:r2r_ce_val_unseen}, Time denotes average task completion time in minutes, and $^{\dagger}$ denotes few-shot evaluation using the Experience Library.

\Needspace{5\baselineskip}
\subsubsection{LIBERO-Pro}
Table~\ref{tab:libero_pro_perturbation} reports results under instruction-redirection (T) and position-swap (S) perturbations across four task suites. \method{} (9B) achieves 65.5\% overall success, improving substantially over the few-shot backbone and larger Qwen3.8-27B. Average task completion time decreases from 13.8 minutes with GPT-5.6-Sol to 1.6 minutes, an $8.6\times$ speedup. RSI further raises success to 69.3\% at the same average completion time, improving robustness without additional execution time.

\begin{table}[!htbp]
\centering
\caption{Success rates (\%) on LIBERO-Pro under Task (T) and Swap (S) perturbations across four suites.}
\label{tab:libero_pro_perturbation}
\small
\setlength{\tabcolsep}{2.5pt}
\renewcommand{\arraystretch}{1.10}
\begin{tabularx}{\linewidth}{l*{10}{>{\centering\arraybackslash}X}}
\toprule
\multirow{2}{*}{Method} & \multicolumn{9}{c}{Success Rate (\%) $\uparrow$} & \multirow{2}{*}{Time $\downarrow$} \\
\cmidrule(lr){2-10}
 & Spat-T & Spat-S & Obj-T & Obj-S & Goal-T & Goal-S & L10-T & L10-S & Overall & \\
\midrule
\rowcolor{tableorange}
\multicolumn{11}{c}{\itshape Training-Based Skill Policies} \\
OpenVLA~\cite{pmlr-v270-kim25c}
  & 0.0 & 0.0 & 0.0 & 0.0 & 0.0 & 0.0 & 0.0 & 0.0 & 0.0 & / \\
NORA \cite{hung2025nora}
  & 0.0 & 0.0 & 0.0 & 0.0 & 0.0 & 0.0 & 0.0 & 0.0 & 0.0 & / \\
$\pi_0$~\cite{BlackK-RSS-25}
  & 0.0 & 0.0 & 0.0 & 2.0 & 0.0 & 0.0 & 0.0 & 0.0 & 0.3 & / \\
MolmoAct \cite{lee2025molmoact}
  & 0.0 & 0.0 & 0.0 & 6.0 & 0.0 & 0.0 & 6.0 & 0.0 & 1.5 & / \\
X-VLA~\cite{zheng2026x}
  & 0.0 & 0.0 & 8.0 & 2.0 & 9.0 & 1.0 & 10.0 & 0.0 & 3.8 & / \\
AtomVLA \cite{sun2026atomvla}
  & 1.0 & 16.0 & 0.0 & 10.0 & 11.0 & 2.0 & 9.0 & 1.0 & 6.3 & / \\
$\pi_{0.5}$~\cite{pmlr-v305-black25a}
  & 41.0 & 52.0 & 56.0 & 69.0 & 42.0 & 36.0 & 46.0 & 8.0 & 43.8 & \textbf{0.6} \\
\midrule
\rowcolor{tableorange}
\multicolumn{11}{c}{\itshape Few-Shot Agents} \\
Qwen3.5-4B$^{\dagger}$
  & 6.0 & 10.0 & 10.0 & 17.0 & 21.0 & 14.0 & 6.0 & 1.0 & 10.6 & 9.5\\
Qwen3.5-9B$^{\dagger}$
  & 6.0 & 11.0 & 28.0 & 13.0 & 13.0 & 20.0 & 22.0 & 12.0 & 15.6 & 6.2 \\
Qwen3.8-27B$^{\dagger}$
  & 13.0 & 18.0 & 55.0 & 48.0 & 24.0 & 17.0 & 24.0 & 9.0 & 26.0 & 9.2 \\
GLM-5.3-Flash$^{\dagger}$
  & 68.0 & 67.0 & 76.0 & 80.0 & 69.0 & \underline{71.0} & \underline{52.0} & \underline{36.0} & 64.9 & 8.5 \\
\textbf{GPT-5.6-Sol}$^{\dagger}$
  & \textbf{92.0} & \textbf{81.0} & \textbf{95.0} & \textbf{97.0}
  & \textbf{84.0} & \textbf{83.0} & \textbf{57.0} & \textbf{37.0}
  & \textbf{78.3} & 13.8 \\
\midrule
\rowcolor{tableorange}
\textbf{\method{} (4B)} & 77.0 & 69.0 & 90.0 & 92.0 & 66.0 & 57.0 & 33.0 & 15.0 & 62.4 & \underline{1.5} \\
\rowcolor{tableorange}
\textbf{\method{} (9B)}
  & 78.0 & 75.0 & \underline{93.0} & \underline{96.0}
  & 64.0 & 52.0 & 44.0 & 22.0 & 65.5 & 1.6 \\
\rowcolor{tableorange}
\textbf{\method{} (RSI)} & \underline{84.0} & \underline{78.0} & 92.0 & 95.0 & \underline{70.0} & 61.0 & 48.0 & 26.0 & \underline{69.3} & 1.6 \\
\bottomrule
\end{tabularx}
\end{table}

\Needspace{5\baselineskip}
\subsubsection{LIBERO-Plus}
Table~\ref{tab:libero_plus} evaluates robustness across seven perturbation dimensions. \method{} (9B) achieves 83.8\% overall success, exceeding all compared skill policies and few-shot GLM-5.3-Flash (69.3\%). Its largest gains over $\pi_{0.5}$ (83.4\%) occur under language and robot-state perturbations. RSI raises overall success to 84.5\%, second only to GPT-5.6-Sol (86.2\%) among the compared methods. The 9B and RSI variants both complete tasks in 2.2 minutes on average, compared with 10.5 minutes for GPT-5.6-Sol, a $4.8\times$ speedup.

\begin{table}[!htbp]
\centering
\caption{Success rates (\%) on LIBERO-Plus across seven perturbation types.}
\label{tab:libero_plus}
\small
\setlength{\tabcolsep}{2.5pt}
\renewcommand{\arraystretch}{1.10}
\begin{tabularx}{\linewidth}{l*{9}{>{\centering\arraybackslash}X}}
\toprule
\multirow{2}{*}{Method} & \multicolumn{8}{c}{Success Rate (\%) $\uparrow$} & \multirow{2}{*}{Time $\downarrow$}\\
\cmidrule(lr){2-9}
 & Camera & Robot & Language & Light & \makecell{Back-\\ground} & Noise & Layout & Overall & \\
\midrule
\rowcolor{tableorange}
\multicolumn{10}{c}{\itshape Training-Based Skill Policies} \\
OpenVLA~\cite{pmlr-v270-kim25c}
  & 0.8 & 3.5 & 23.0 & 8.1 & 34.8 & 15.2 & 28.5 & 15.6 & / \\
NORA~\cite{hung2025nora}
  & 2.2 & 37.0 & 65.1 & 45.7 & 58.6 & 12.8 & 62.1 & 39.0 & / \\
$\pi_0$~\cite{BlackK-RSS-25}
  & 13.8 & 6.0 & 58.8 & 85.0 & 81.4 & 79.0 & 68.9 & 53.6 & / \\
OpenVLA-OFT~\cite{KimM-RSS-25}
  & 56.4 & 31.9 & 79.5 & 88.7 & 93.3 & 75.8 & 74.2 & 69.6 & / \\
ABot-M0~\cite{yang2026abotm0}
  & 60.4 & 67.9 & \underline{86.4} & \textbf{96.2} & 91.6 & \textbf{86.4} & 82.6 & 80.5 & / \\
$\pi_{0.5}$~\cite{pmlr-v305-black25a}
  & 69.2 & 74.2 & 83.3 & 94.2 & 95.0 & \underline{83.3} & \underline{85.0} & 83.4 & \textbf{1.3} \\
\midrule
\rowcolor{tableorange}
\multicolumn{10}{c}{\itshape Few-Shot Agents} \\
Qwen3.5-4B$^{\dagger}$
  & 12.5 & 12.5 & 13.3 & 10.8 & 10.0 & 8.3 & 5.8 & 10.5 & 3.5 \\
Qwen3.5-9B$^{\dagger}$
  & 12.5 & 15.8 & 15.8 & 10.8 & 15.0 & 13.3 & 16.7 & 14.3 & \underline{2.1} \\
Qwen3.8-27B$^{\dagger}$
  & 25.0 & 25.8 & 39.2 & 31.7 & 27.5 & 28.3 & 27.5 & 29.3 & 4.5 \\
GLM-5.3-Flash$^{\dagger}$
  & 60.0 & 56.7 & 77.5 & 74.2 & 71.7 & 73.3 & 71.7 & 69.3 & 7.8 \\
\textbf{GPT-5.6-Sol}$^{\dagger}$
  & \textbf{72.5} & \textbf{79.2} & \textbf{88.3} & \underline{95.0}
  & \textbf{98.3} & \underline{83.3} & \textbf{86.7} & \textbf{86.2} & 10.5 \\
\midrule
\rowcolor{tableorange}
\textbf{\method{} (4B)}
  & \underline{70.8} & 73.3 & 84.2 & 90.0 & 93.3 & 81.7 & 82.5 & 82.3 & \underline{2.1} \\
\rowcolor{tableorange}
\textbf{\method{} (9B)}
  & 70.0 & 75.8 & 85.8 & 92.5 & 95.8 & 82.5 & 84.2 & 83.8 & 2.2 \\
\rowcolor{tableorange}
\textbf{\method{} (RSI)}
  & \underline{70.8} & \underline{76.7} & 85.0 & 94.2 & \underline{96.7} & \underline{83.3}
  & \underline{85.0} & \underline{84.5} & 2.2 \\
\bottomrule
\end{tabularx}
\end{table}

\Needspace{5\baselineskip}
\subsubsection{RoboCasa365}
Table~\ref{tab:robocasa365} reports results on Atomic-Seen, Composite-Seen, and Composite-Unseen tasks. GPT-5.6-Sol achieves the highest success rates across all splits, while \method{} (9B) reaches 39.6\% overall success, exceeding RLDX-1 (36.0\%) and more than tripling the few-shot Qwen3.5-9B result (12.2\%). Average task completion time is 13.3 minutes, compared with 14.8 minutes for GPT-5.6-Sol and 21.3 minutes for the few-shot backbone, corresponding to reductions of 10.1\% and 37.6\%, respectively. RSI further improves success to 43.2\% with a slightly shorter completion time of 13.0 minutes. Unseen task compositions remain the largest gap to the teacher.

\begin{table}[!htbp]
\centering
\caption{Success rates (\%) on RoboCasa365 across seen and unseen task splits.}
\label{tab:robocasa365}
\small
\setlength{\tabcolsep}{5pt}
\renewcommand{\arraystretch}{1.10}
\begin{tabularx}{\linewidth}{l*{5}{>{\centering\arraybackslash}X}}
\toprule
\multirow{2}{*}{Method}
  & \multicolumn{4}{c}{Success Rate (\%) $\uparrow$}
  & \multirow{2}{*}{Time $\downarrow$} \\
\cmidrule(lr){2-5}
 & Atomic-Seen & Composite-Seen & Composite-Unseen & Overall & \\
\midrule
\rowcolor{tableorange}
\multicolumn{6}{c}{\itshape Training-Based Skill Policies} \\
Diffusion Policy~\cite{chi2024diffusionpolicy}
  & 15.7 & 0.2 & 1.3 & 6.1 & / \\
$\pi_0$~\cite{BlackK-RSS-25}
  & 34.6 & 6.1 & 1.1 & 14.8 & / \\
$\pi_{0.5}$~\cite{pmlr-v305-black25a}
  & 39.6 & 7.1 & 1.2 & 16.9 & / \\
Giga World-Policy 0.1~\cite{ye2026gigaworldpolicy}
  & 44.4 & 11.8 & 2.9 & 20.7 & / \\
GR00T N1.5~\cite{gr00tn1_2025}
  & 50.7 & 14.8 & 2.7 & 23.9 & / \\
RLDX-1~\cite{rldx2026}
  & 67.6 & 27.9 & 8.5 & 36.0 & \textbf{5.3} \\
\midrule
\rowcolor{tableorange}
\multicolumn{6}{c}{\itshape Few-Shot Agents} \\
Qwen3.5-4B$^{\dagger}$
  & 27.2 & 6.3 & 0.0 & 11.8 & 14.5 \\
Qwen3.5-9B$^{\dagger}$
  & 28.3 & 6.3 & 0.0 & 12.2 & 21.3 \\
Qwen3.8-27B$^{\dagger}$
  & 37.2 & 11.3 & 2.5 & 17.8 & 29.8 \\
GLM-5.3-Flash$^{\dagger}$
  & \underline{74.4} & 28.8 & 13.8 & 40.4 & 16.5 \\
\textbf{GPT-5.6-Sol}$^{\dagger}$
  & \textbf{81.1} & \textbf{45.0} & \textbf{27.5}
  & \textbf{52.4} & 14.8 \\
\midrule
\rowcolor{tableorange}
\textbf{\method{} (4B)}
  & 60.0 & 30.0 & 7.5 & 33.6 & \underline{8.9} \\
\rowcolor{tableorange}
\textbf{\method{} (9B)}
  & 71.1 & 32.5 & 11.3 & 39.6 & 13.3 \\
\rowcolor{tableorange}
\textbf{\method{} (RSI)}
  & \underline{74.4} & \underline{36.3}
  & \underline{15.0} & \underline{43.2} & 13.0 \\
\bottomrule
\end{tabularx}
\end{table}

\Needspace{5\baselineskip}
\subsubsection{RoboTwin 2.0}
Table~\ref{tab:robotwin} reports success rates under Clean and Randomized configurations. \method{} (9B) achieves 88.8\% and 87.6\%, respectively, with an overall success rate of 88.2\%, exceeding LingBot-VLA and few-shot GLM-5.3-Flash. Average task completion time is 3.7 minutes, compared with 10.5 minutes for GPT-5.6-Sol, a $2.8\times$ speedup with a 3.0-point gap in overall success. RSI narrows this gap to 1.8 points without increasing average completion time.

\begin{table}[!htbp]
\centering
\caption{Success rates (\%) on RoboTwin~2.0 under Clean and Randomized settings.}
\label{tab:robotwin}
\small
\setlength{\tabcolsep}{5pt}
\renewcommand{\arraystretch}{1.10}
\begin{tabularx}{\linewidth}{l*{4}{>{\centering\arraybackslash}X}}
\toprule
\multirow{2}{*}{Method}
  & \multicolumn{3}{c}{Success Rate (\%) $\uparrow$}
  & \multirow{2}{*}{Time $\downarrow$} \\
\cmidrule(lr){2-4}
 & Clean & Randomized & Overall & \\
\midrule
\rowcolor{tableorange}
\multicolumn{5}{c}{\itshape Training-Based Skill Policies} \\
X-VLA~\cite{zheng2026x}
  & 72.8 & 41.2 & 57.0 & / \\
$\pi_{0.5}$~\cite{pmlr-v305-black25a}
  & 81.2 & 75.6 & 78.4 & / \\
LingBot-VLA~\cite{wu2026pragmatic}
  & 86.0 & 83.6 & 84.8 & \textbf{2.8} \\
ABot-M0~\cite{yang2026abotm0}
  & 86.06 & 85.08 & 85.57 & / \\
Qwen-VLA~\cite{wang2026qwenvla}
  & 86.10 & 87.20 & 86.65 & / \\
\midrule
\rowcolor{tableorange}
\multicolumn{5}{c}{\itshape Few-Shot Agents} \\
Qwen3.5-4B$^{\dagger}$
  & 59.2 & 62.4 & 60.8 & 4.4 \\
Qwen3.5-9B$^{\dagger}$
  & 62.8 & 60.4 & 61.6 & 5.0 \\
Qwen3.8-27B$^{\dagger}$
  & 68.4 & 67.6 & 68.0 & 8.7 \\
GLM-5.3-Flash$^{\dagger}$
  & 87.6 & 86.4 & 87.0 & 9.1 \\
\textbf{GPT-5.6-Sol}$^{\dagger}$
  & \textbf{92.0} & \textbf{90.4} & \textbf{91.2} & 10.5 \\
\midrule
\rowcolor{tableorange}
\textbf{\method{} (4B)}
  & 88.0 & 86.4 & 87.2 & \underline{3.2} \\
\rowcolor{tableorange}
\textbf{\method{} (9B)}
  & 88.8 & 87.6 & 88.2 & 3.7 \\
\rowcolor{tableorange}
\textbf{\method{} (RSI)}
  & \underline{90.4} & \underline{88.4} & \underline{89.4} & 3.7 \\
\bottomrule
\end{tabularx}
\end{table}

\Needspace{5\baselineskip}
\subsubsection{R2R-CE}
Table~\ref{tab:r2r_ce_val_unseen} reports navigation results; comparisons among agents use the same 100-episode \texttt{rand100} subset of Val-Unseen. Relative to the few-shot Qwen3.5-9B backbone, \method{} (9B) improves SR from 34.0\% to 65.0\% and SPL from 26.9\% to 54.8\%, while reducing NE from 6.57\,m to 3.78\,m. It also exceeds few-shot Qwen3.8-27B and GLM-5.3-Flash on these metrics. The remaining gap to GPT-5.6-Sol is 6.0 points in SR and 2.7 points in SPL, showing that much of the navigation capability is retained after distillation. An OSR of 71.0\% versus 65.0\% SR also suggests room to improve endpoint selection and termination. Average task completion time decreases from 6.03 minutes with GPT-5.6-Sol to 0.86 minutes, a $7.0\times$ speedup. RSI improves SR to 66.0\% with a similar completion time of 0.87 minutes.

\begin{table}[!htbp]
\centering
\caption{R2R-CE Val-Unseen results. Skill policies use the full split; agents use the 100-episode \texttt{rand100} subset. SR, SPL, and OSR are percentages; NE is in meters.}
\label{tab:r2r_ce_val_unseen}
\small
\setlength{\tabcolsep}{5pt}
\renewcommand{\arraystretch}{1.10}
\begin{tabularx}{\linewidth}{l*{5}{>{\centering\arraybackslash}X}}
\toprule
Method & SR $\uparrow$ & SPL $\uparrow$ & OSR $\uparrow$ & NE $\downarrow$ & Time $\downarrow$ \\
\midrule
\rowcolor{tableorange}
\multicolumn{6}{c}{\itshape Training-Based Skill Policies} \\
NaVid \cite{Zhang-RSS-24} & 37.4 & 35.9 & 49.1 & 5.47 & / \\
Uni-NaVid~\cite{ZhangJ-RSS-25} & 47.0 & 42.7 & 53.3 & 5.58 & / \\
AO-Planner \cite{chen2025affordances} & 47.0 & 33.0 & 59.0 & 5.55 & / \\
ActiveVLN \cite{zhang2026activevln} & 52.9 & 45.7 & 62.1 & 5.34 & / \\
StreamVLN~\cite{wei2026streamvlnstreamingvisionandlanguagenavigation}
  & 56.9 & 51.9 & \underline{64.2} & 4.98 & / \\ %
Progress-Think \cite{wang2026progressthink} & \underline{60.1} & \underline{53.6} & 63.6 & \underline{4.68} & \textbf{0.61} \\
DualVLN~\cite{wei2026ground} & \textbf{64.3} & \textbf{58.5} & \textbf{70.7} & \textbf{4.05} & / \\
\midrule
\rowcolor{tableorange}
\multicolumn{6}{c}{\itshape Few-Shot Agents} \\
Qwen3.5-4B$^{\dagger}$ & 18.0 & 13.2 & 41.0 & 7.74 & \textbf{0.51} \\
Qwen3.5-9B$^{\dagger}$ & 34.0 & 26.9 & 48.0 & 6.57 & 1.03 \\
Qwen3.8-27B$^{\dagger}$ & 41.0 & 31.6 & 58.0 & 6.12 & 3.26 \\
GLM-5.3-Flash$^{\dagger}$ & 58.0 & 35.5 & \underline{73.0} & 4.31 & 6.46 \\
\textbf{GPT-5.6-Sol}$^{\dagger}$ & \textbf{71.0} & \textbf{57.5} & \textbf{75.0} & \textbf{3.11} & 6.03 \\
\midrule
\rowcolor{tableorange}
\textbf{\method{} (4B)} & 55.0 & 45.9 & 58.0 & 5.01 & \underline{0.72} \\
\rowcolor{tableorange}
\textbf{\method{} (9B)} & 65.0 & 54.8 & 71.0 & 3.78 & 0.86 \\
\rowcolor{tableorange}
\textbf{\method{} (RSI)} & \underline{66.0} & \underline{55.2} & \underline{73.0} & \underline{3.70} & 0.87 \\
\bottomrule
\end{tabularx}
\end{table}

\Needspace{5\baselineskip}
\subsubsection{Spatial understanding and referring}
Table~\ref{tab:embspatial} evaluates egocentric spatial understanding through multiple-choice questions. \method{} (9B) achieves 81.6\% overall accuracy, improving over its Qwen3.5-9B backbone by 4.4 points and obtaining the highest accuracy across all three source environments. These consistent gains suggest that the improvement is not confined to a single scene source.

\begin{table}[!htbp]
\centering
\caption{Spatial question-answering accuracy (\%) on EmbSpatial-Bench across three source environments.}
\label{tab:embspatial}
\small
\setlength{\tabcolsep}{5pt}
\renewcommand{\arraystretch}{1.16}
\begin{tabularx}{\linewidth}{l*{4}{>{\centering\arraybackslash}X}}
\toprule
\multirow{2}{*}{Method} & \multicolumn{4}{c}{Accuracy (\%) $\uparrow$} \\
\cmidrule(lr){2-5}
& MP3D & AI2-THOR & ScanNet & Overall \\
\midrule
InternVL3-8B~\cite{zhu2025internvl3} & 76.3 & \underline{74.2} & 78.5 & 76.3 \\
Cambrian-S-7B~\cite{yang2026cambrian} & 71.3 & 72.9 & 74.3 & 72.8 \\
\midrule
Qwen3.5-4B & 77.7 & 71.8 & 77.1 & 75.5 \\
Qwen3.5-9B & 78.6 & 72.1 & 81.2 & 77.2 \\
\midrule
\rowcolor{tableorange}
\textbf{\method{} (4B)} & \underline{81.1} & 73.0 & \underline{83.2} & \underline{79.0} \\
\rowcolor{tableorange}
\textbf{\method{} (9B)} & \textbf{84.0} & \textbf{76.8} & \textbf{84.3} & \textbf{81.6} \\
\bottomrule
\end{tabularx}
\end{table}

\Needspace{5\baselineskip}
Table~\ref{tab:refspatial} evaluates point prediction for object localization and placement. \method{} (9B) improves Placement and Unseen accuracy over its backbone by 7.0 and 5.2 points, respectively, matching RoboRefer-2B-RFT on Unseen combinations. Its lower Location accuracy leaves overall performance slightly below the backbone, indicating uneven transfer across spatial referring tasks. This pattern suggests stronger transfer to relational placement and compositional queries than to direct object localization.

\begin{table}[!htbp]
\centering
\caption{Spatial referring accuracy (\%) on RefSpatial-Bench: object localization (Location), free-space selection (Placement), and novel relation combinations (Unseen).}
\label{tab:refspatial}
\small
\setlength{\tabcolsep}{5pt}
\renewcommand{\arraystretch}{1.16}
\begin{tabularx}{\linewidth}{l*{4}{>{\centering\arraybackslash}X}}
\toprule
\multirow{2}{*}{Method} & \multicolumn{4}{c}{Accuracy (\%) $\uparrow$} \\
\cmidrule(lr){2-5}
& Location & Placement & Unseen & Overall \\
\midrule
SpaceLLaVA~\cite{remyx2024spacellava} & 5.8 & 4.3 & 4.0 & 4.8 \\
RoboPoint~\cite{yuan2024robopoint} & 22.9 & 9.3 & 8.4 & 13.9 \\
Molmo-7B~\cite{deitke2025molmo} & 21.9 & 12.9 & 12.2 & 15.9 \\
RoboRefer-2B-RFT~\cite{zhou2026roborefer} & \underline{52.0} & \textbf{54.0} & \textbf{41.6} & \textbf{49.8} \\
\midrule
Qwen3.5-4B & 46.0 & 20.0 & 26.0 & 31.1 \\
Qwen3.5-9B & \textbf{65.0} & 41.0 & 36.4 & \underline{48.4} \\
\midrule
\rowcolor{tableorange}
\textbf{\method{} (4B)} & 43.0 & 46.0 & \underline{37.7} & 42.6 \\
\rowcolor{tableorange}
\textbf{\method{} (9B)} & \underline{52.0} & \underline{48.0} & \textbf{41.6} & 47.7 \\
\bottomrule
\end{tabularx}
\end{table}

\FloatBarrier
\begin{figure}[H]
\noindent
\begin{minipage}[t]{0.58\linewidth}
  \vspace{0pt}
  \subsubsection{Real-world manipulation}
  Figure~\ref{fig:realworld} summarizes results on three real-world
  manipulation tasks: block stacking, apple placement, and drawer opening.
  OmniBrain-0 achieves a macro-average success rate of 73.3\%,
  outperforming $\pi_{0.5}$, which achieves 58.3\% across the three tasks.
  The improvement is primarily driven by block stacking, where success
  increases from 50\% to 75\%, followed by apple placement (40\% to 60\%),
  while performance on drawer opening is matched at 85\%.
  GPT-5.6-Sol~(few-shot) achieves a higher macro-average of 86.7\%,
  with the largest gaps to \method{} on block stacking
  (90\% vs.\ 75\%) and apple placement (75\% vs.\ 60\%).
  These results support the applicability of the compact unified agent
  to the evaluated physical manipulation tasks, although further
  improvements are needed to match the teacher's task-level performance.
\end{minipage}\hfill
\begin{minipage}[t]{0.39\linewidth}
  \vspace{0pt}
  \setlength{\abovecaptionskip}{3pt}
  \centering
  \includegraphics[width=\linewidth]{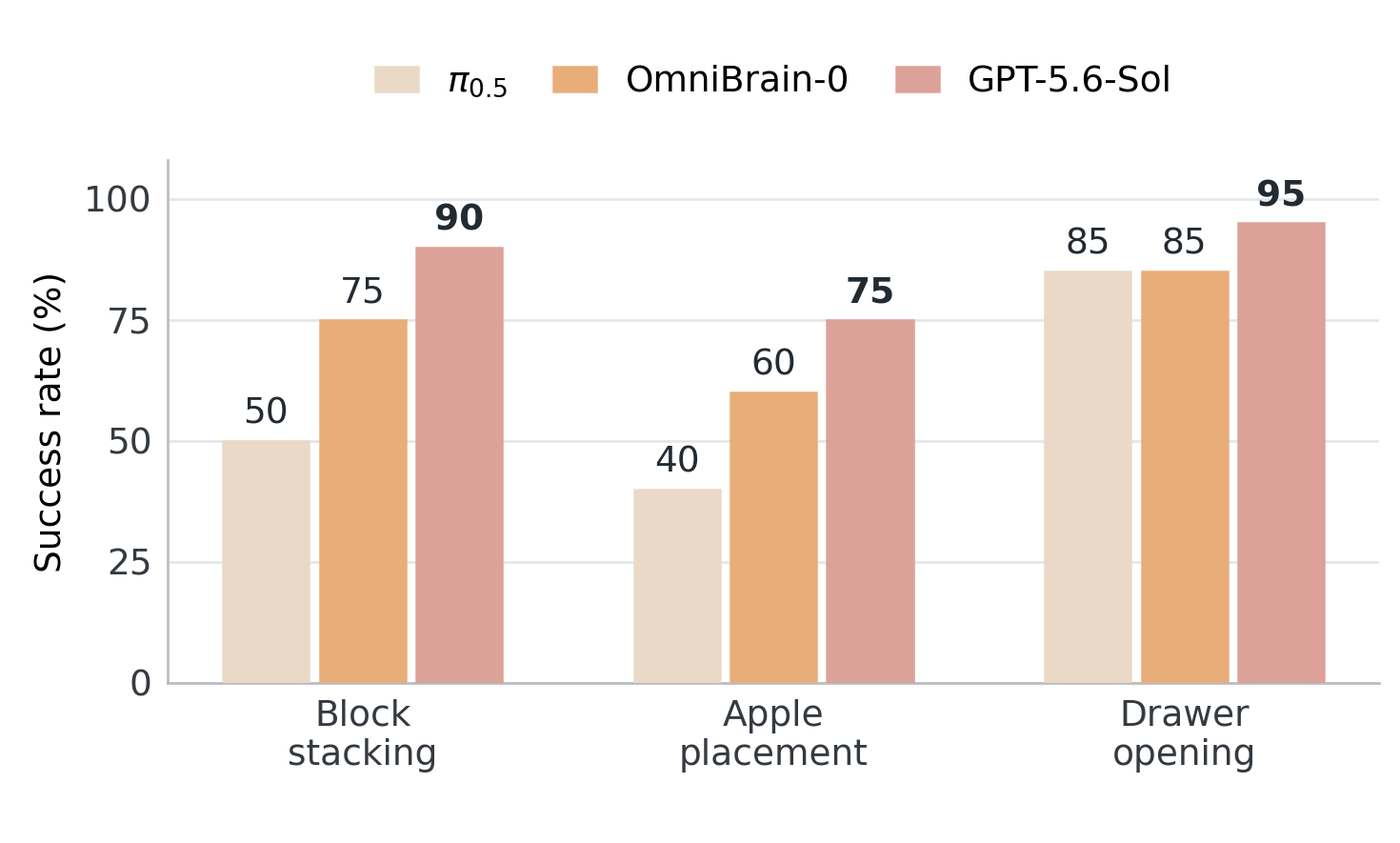}
  \caption{\textbf{Real-world manipulation performance.}
  Comparison of the standalone baseline $\pi_{0.5}$,
  the compact unified agent \method{}, and
  GPT-5.6-Sol~(few-shot). Success rates are reported as percentages.}
  \label{fig:realworld}
\end{minipage}
\end{figure}

\FloatBarrier
\Needspace{5\baselineskip}
\subsubsection{Inference Token Efficiency}
Table~\ref{tab:token_efficiency} reports average token consumption per task. Relative to GPT-5.6-Sol, \method{} (9B) uses $4.5\times$, $2.6\times$, $3.5\times$, $12.7\times$, and $3.7\times$ fewer tokens on LIBERO-Pro, LIBERO-Plus, RoboCasa365, RoboTwin~2.0, and R2R-CE, respectively. These token savings accompany the shorter task completion times reported above, with speedups ranging from $1.1\times$ to $8.6\times$ across the five benchmarks. The gains differ across tasks, since end-to-end completion time also includes tool calls and low-level execution.

\begin{table}[!htbp]
\centering
\caption{Average token consumption per task (millions of tokens, M).}
\label{tab:token_efficiency}
\small
\setlength{\tabcolsep}{3pt}
\renewcommand{\arraystretch}{1.16}
\begin{tabularx}{\linewidth}{l*{5}{>{\centering\arraybackslash}X}}
\toprule
Method & LIBERO-Pro & LIBERO-Plus & RoboCasa365 & RoboTwin 2.0 & R2R-CE \\
\midrule
GPT-5.6-Sol & 1.7 & 2.2 & 3.0 & 1.4 & 0.163 \\
\midrule
\textbf{\method{} (4B)} &\underline{0.65} & \underline{1.2} & \textbf{0.71} & \underline{0.13} & \textbf{0.042}  \\

\rowcolor{tableorange}
\textbf{\method{} (9B)}& \textbf{0.38} & \textbf{0.86} & \underline{0.85} & \textbf{0.11} & \underline{0.044}\\
\bottomrule
\end{tabularx}
\end{table}

\FloatBarrier

\subsection{Ablation Studies}
\label{sec:ablations}
\label{appendx:ablation}

\begin{figure}[!htbp]
    \centering
    \includegraphics[width=0.96\linewidth]{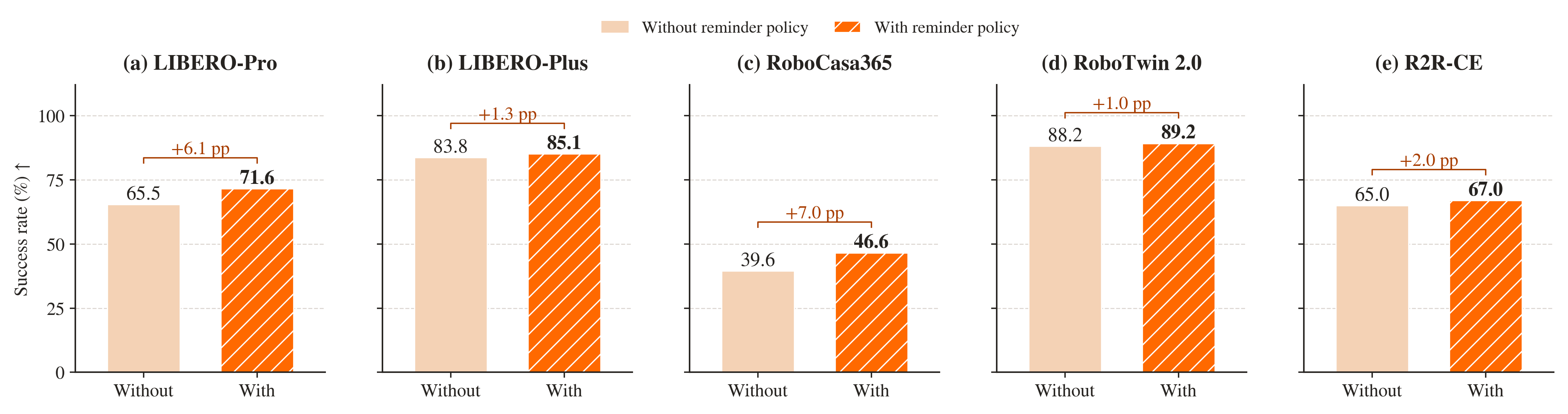}
    \caption{\textbf{Inference-time reminder-policy ablation.} Success rates for execution without reminders and with the state-dependent reminder policy. The panels report Overall for LIBERO-Pro, LIBERO-Plus, RoboCasa365, and RoboTwin~2.0, and SR for R2R-CE.}
    \label{fig:ablation_reminder_policy}
\end{figure}

\paragraph{Ablation Study on the Reminder Policy}
Figure~\ref{fig:ablation_reminder_policy} isolates the inference-time contribution of the reminder policy refined through the iterations in Section~\ref{sec:reminder_policy_iteration}. With the model checkpoint fixed, we compare execution without reminders and with state-dependent reminders conditionally inserted before each high-level decision when relevant issues are detected. The comparison uses the same task splits, seeds, tool interfaces, and execution budgets for both conditions, measuring whether the iteratively refined guidance improves task success independently of model-parameter updates.

\FloatBarrier

\begin{figure}[!htbp]
    \centering
    \includegraphics[width=0.96\linewidth]{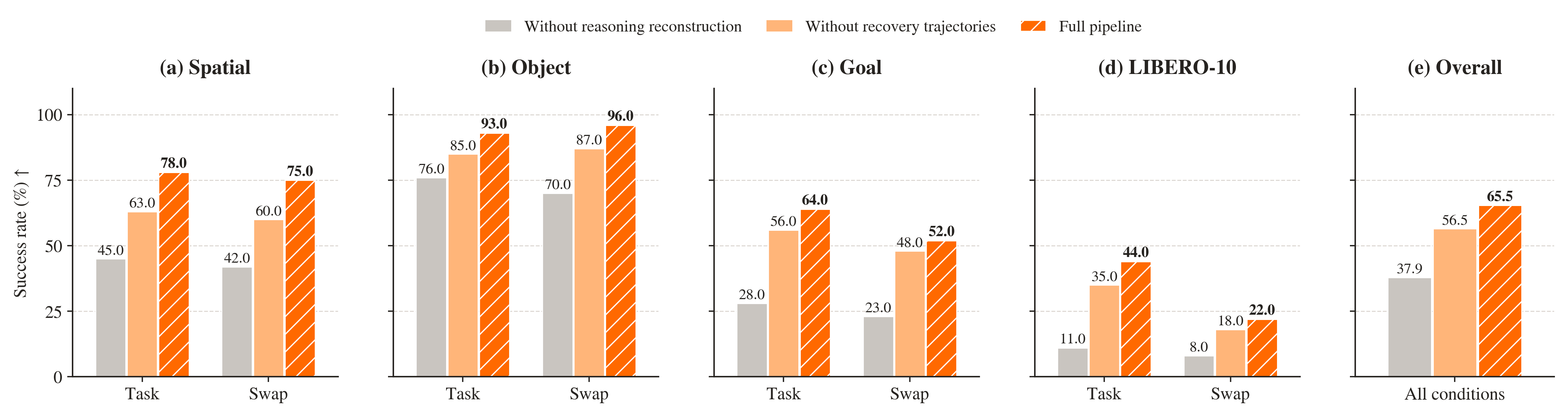}
    \caption{\textbf{Data-pipeline ablation on LIBERO-Pro.} Success rates for removing reasoning reconstruction, removing recovery trajectories, and using the full data pipeline across task and swap conditions.}
    \label{fig:ablation_data_pipeline}
\end{figure}

\paragraph{Ablation Study on the Data Processing Pipeline}
The proposed data processing pipeline is designed to transform expert interaction data into a format suitable for training compact models. The ablation results on LIBERO-Pro in Figure~\ref{fig:ablation_data_pipeline} demonstrate the impact of key components of the pipeline. When the final training data excludes expert failure-recovery trajectories and retains only a single successful trajectory, the model trained on this ablated dataset suffers an 9\% drop in success rate. This degradation highlights the value of failure-recovery trajectories, which provide supervision for actively selecting corrective actions in response to concrete failure evidence, such as failed grasps or object drops during transportation. Furthermore, when the expert-generated reasoning is used directly as free-form text rather than being structurally rewritten, the model trained on the resulting data also exhibits a substantial performance drop. This is because expert models typically produce lengthy and unconstrained reasoning, and directly using such reasoning to construct long-horizon training trajectories for compact models can lead to overfitting to the linguistic style of the expert model, while weakening the learning of visual grounding and decision-making logic.

\FloatBarrier

\begin{figure}[!htbp]
    \centering
    \includegraphics[width=0.96\linewidth]{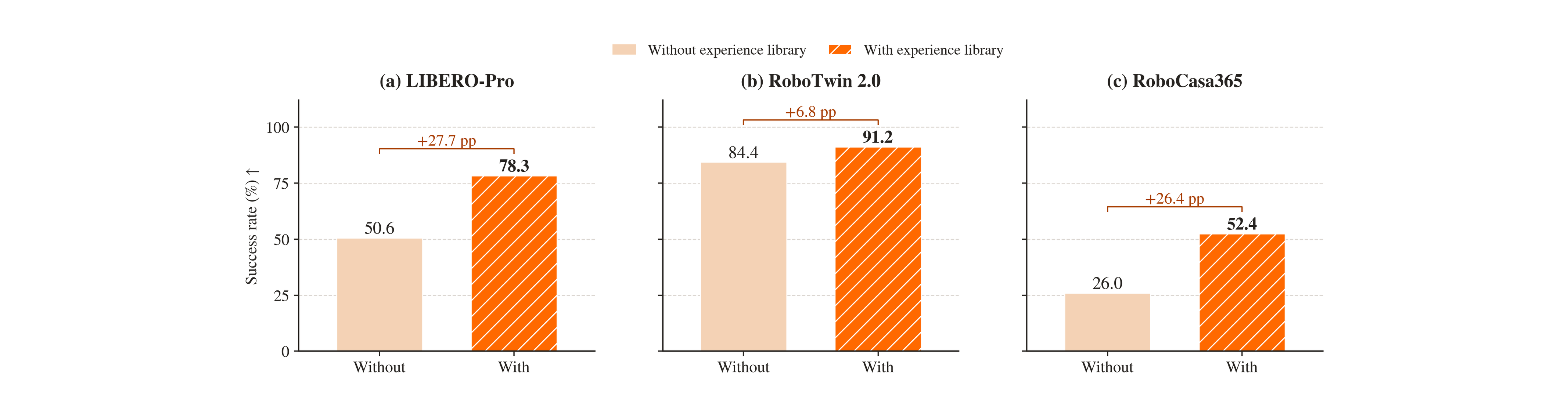}
    \caption{\textbf{Experience-library ablation.} Success rates with and without access to the Experience Library on LIBERO-Pro, RoboTwin~2.0, and RoboCasa365.}
    \label{fig:ablation_experience_library}
\end{figure}

\paragraph{Ablation Study on Experience Library}
Figure~\ref{fig:ablation_experience_library} evaluates the effect of the Experience Library on agent execution, using GPT-5.6-Sol with and without access to the library. The library consistently improves the success rate across all three benchmarks, with gains of 27.7, 6.8, and 26.4 points on LIBERO-Pro, RoboTwin 2.0, and RoboCasa365, respectively. By providing reusable priors from previously collected cross-episode experiences, the Experience Library helps the agent avoid repeated exploration and make more informed execution decisions. Since only successful trajectories are retained for training, the Experience Library also increases the yield of canonical and recovery trajectories available for distillation into \method{}.

\FloatBarrier

\FloatBarrier

\begingroup
\setlength{\intextsep}{4pt plus 1pt minus 1pt}
\setlength{\parskip}{3.5pt plus 1pt minus 1pt}
\setlength{\textfloatsep}{8pt plus 1pt minus 1pt}
\subsection{Qualitative Analysis}
\label{sec:qualitative_analysis}
\label{app:visualization}

Figures~\ref{fig:main-vis-libero}, \ref{fig:main-vis-robotwin}, \ref{fig:main-vis-robocasa}, and~\ref{fig:main-vis-r2r} present representative executions on LIBERO-Pro, RoboTwin~2.0, RoboCasa365, and R2R-CE, respectively, while Figure~\ref{fig:real-world} shows physical robot execution. The examples pair observations with reasoning and tool calls to illustrate object grounding, skill composition, progress tracking, and feedback-driven recovery. Figures~\ref{fig:rsi_vis} and~\ref{fig:rsi_vis_vln} then examine when state-dependent reminders guide manipulation and navigation. Complete manipulation traces, navigation execution examples, and comparisons are provided in Appendix~\ref{appendix:qualitative}.

\subsubsection{Task Execution Process}
\label{sec:qualitative_task_execution}

\paragraph{LIBERO-Pro}
Figure~\ref{fig:main-vis-libero} presents stove operation and book placement. In the stove task (top), the initial grasp fails and the robot reaches the stove without the moka pot (stages 2--3). The agent responds by turning on the stove, refining the grasp cue, re-grasping the pot, and transporting it onto the burner before verifying completion. In the book-placement task (bottom), the agent localizes the book and the caddy, secures and lifts the book, aligns it with the back compartment, and releases it before checking the goal. The sequence shows how perception, VLA skills, and explicit arm motions are coordinated through a shared task state.

\begin{figure}[!htb]
\centering
\includegraphics[width=\textwidth]{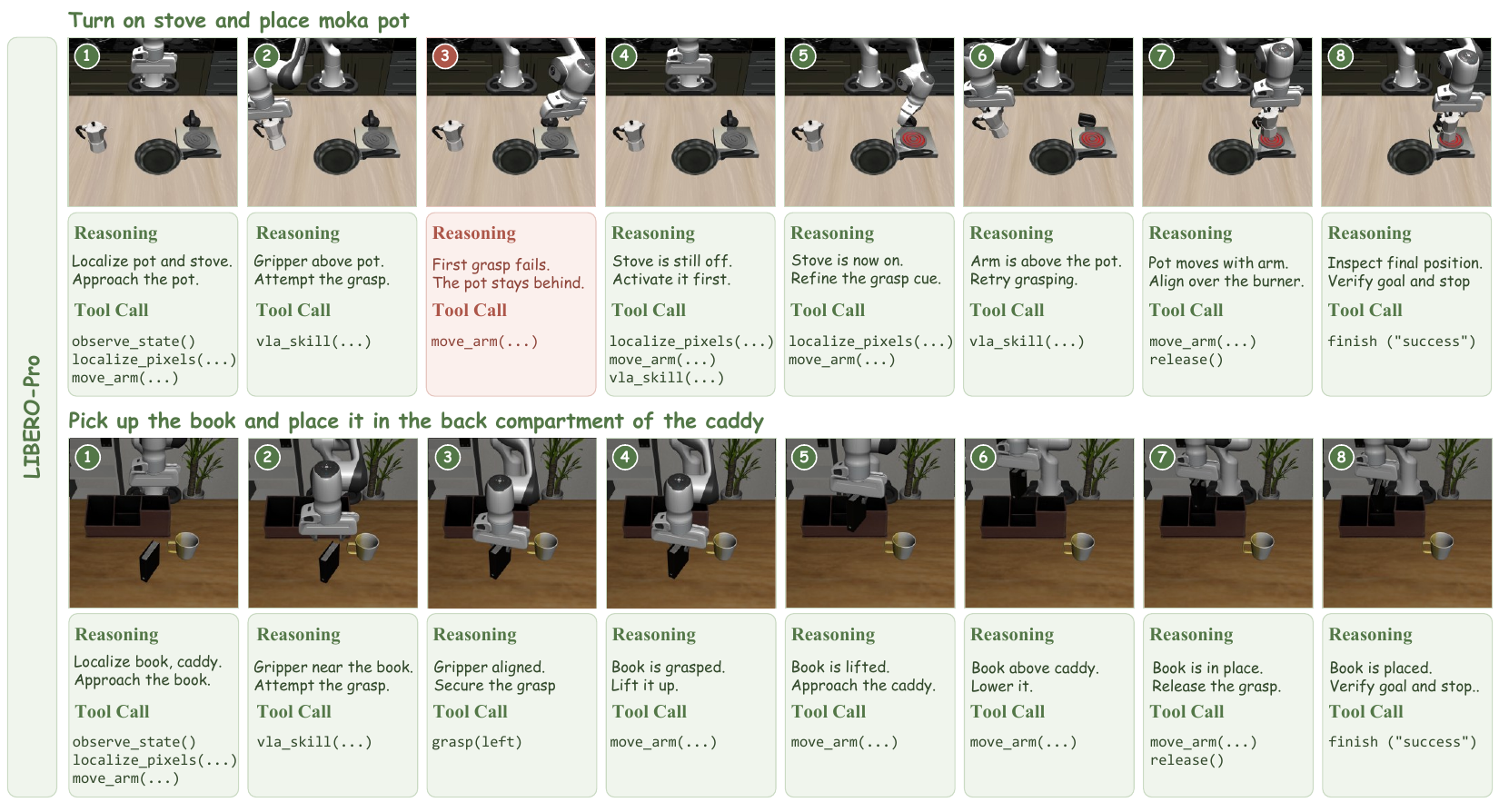}
\caption{\textbf{Examples of \method{} on LIBERO-Pro.} Top: turning on the stove and placing a moka pot on it. Bottom: placing a book in the back compartment of a caddy. Red annotations highlight the failed grasp.}
\label{fig:main-vis-libero}
\end{figure}

\paragraph{RoboTwin 2.0}
Figure~\ref{fig:main-vis-robotwin} examines sequential object placement and articulated-object interaction. In the colored-block task, the agent tracks which blocks have been grasped and placed as it arranges red, green, and blue blocks in a row. Intermediate decisions maintain grasp state during transport and align each new block with those already placed. The microwave task requires a different interaction pattern: the agent localizes the handle, establishes a grasp, and follows the door's swing while maintaining contact and clearance. These cases illustrate how the high-level model selects and coordinates skills according to the current object state, using placement verification for one task and continuous handle interaction for the other.

\begin{figure}[!htb]
\centering
\includegraphics[width=\textwidth]{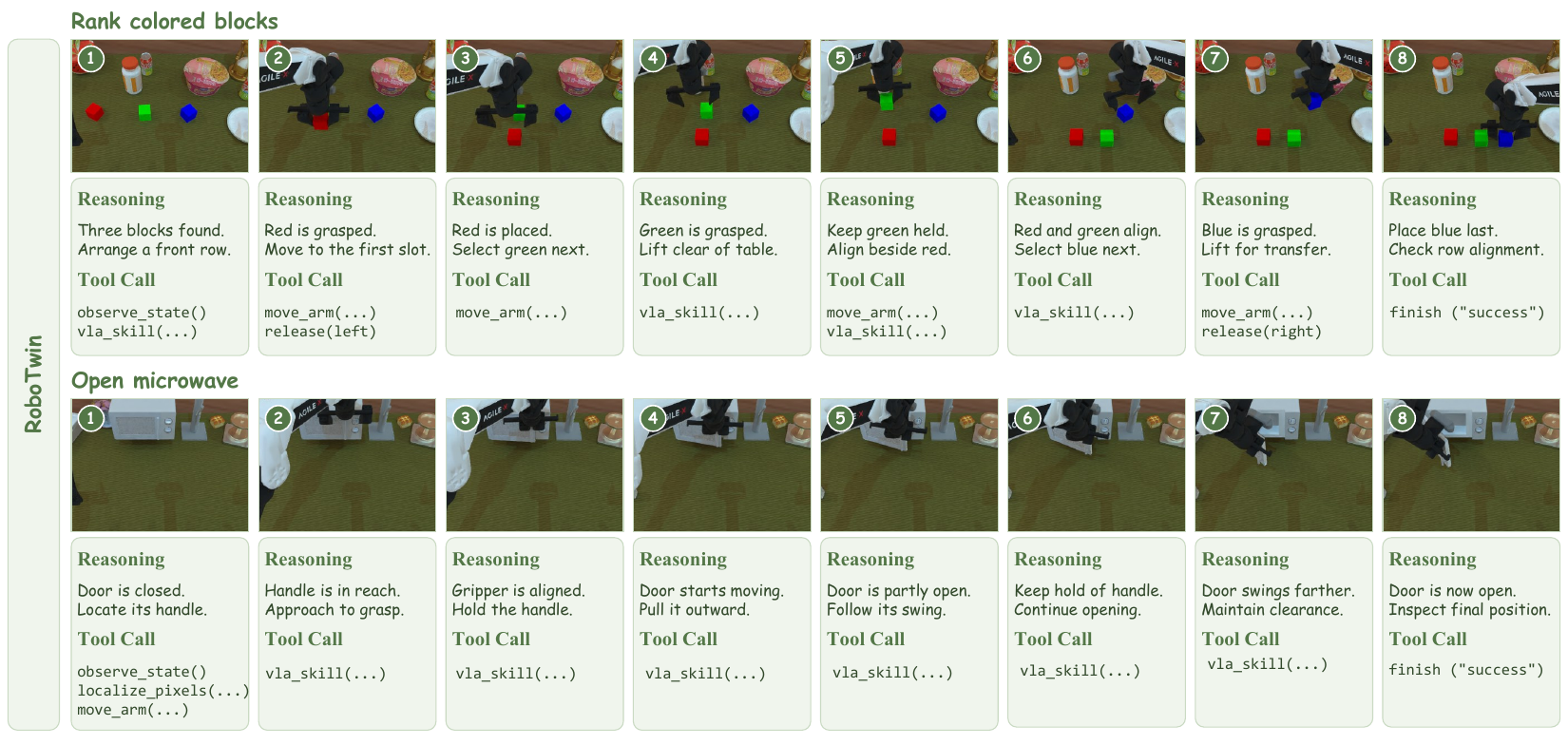}
\caption{\textbf{Examples of \method{} on RoboTwin~2.0.} Top: arranging colored blocks in a row. Bottom: opening a microwave.}
\label{fig:main-vis-robotwin}
\end{figure}

\paragraph{RoboCasa365}
Figure~\ref{fig:main-vis-robocasa} shows household tasks with dependent sub-goals and changes in manipulation context. In the vegetable task (top), an off-target pepper prompts renewed localization and a grasp retry before the agent places it in the pot, handles the next vegetable, and introduces the utensil. The correction follows the observed object state rather than repeating the same action unchanged. In the bread-basket task (bottom), the agent first gains access through the cabinet, grasps and transfers the bread into the basket, and then moves the basket to the island. Tracking these intermediate outcomes allows it to progress from access and object handling to transport of the assembled result.

\begin{figure}[!htb]
\centering
\includegraphics[width=\textwidth]{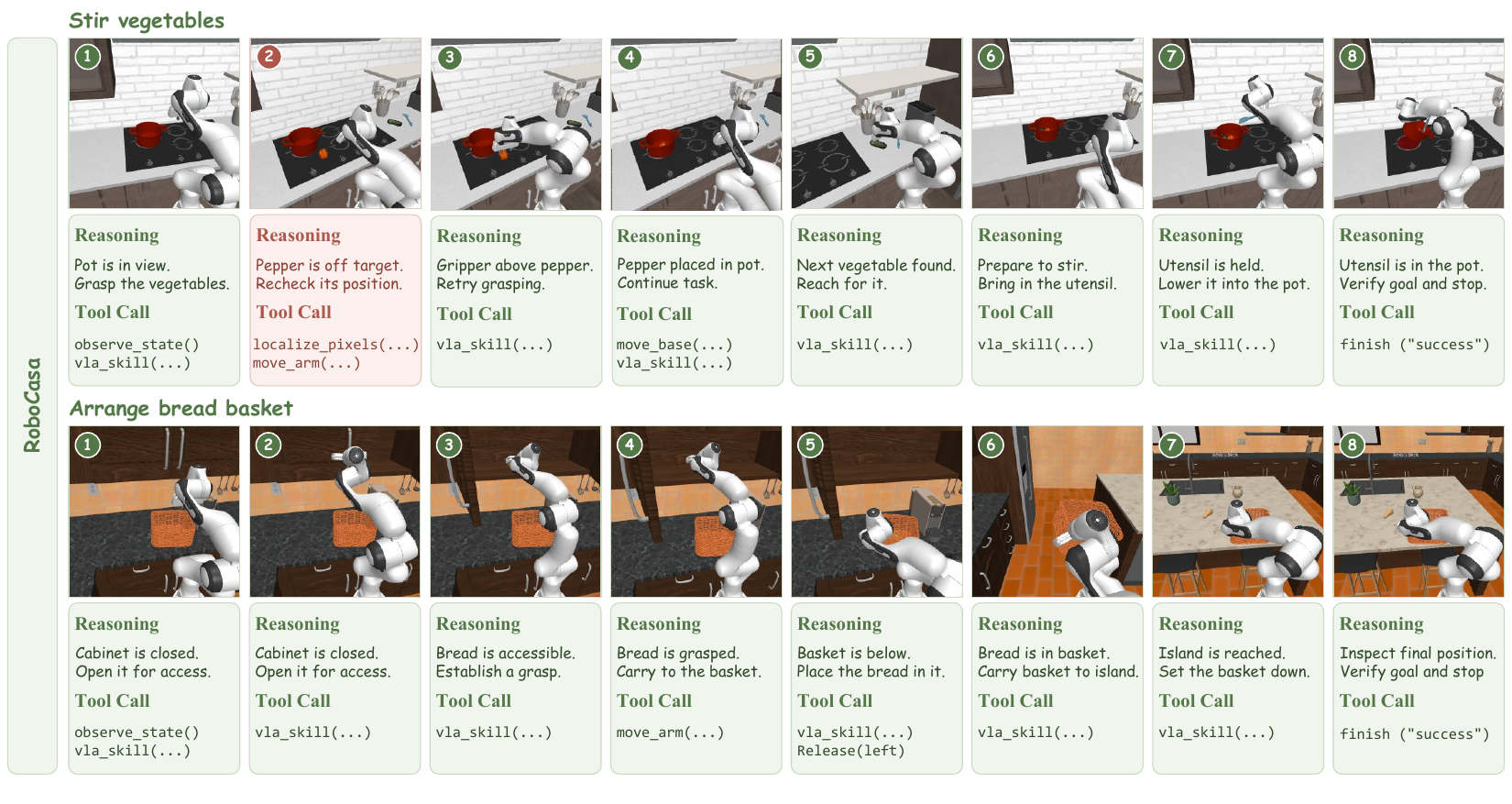}
\caption{\textbf{Examples of \method{} on RoboCasa365.} Top: stirring vegetables. Bottom: arranging a bread basket. Red annotations highlight a failed attempt in the vegetable task.}
\label{fig:main-vis-robocasa}
\end{figure}

\paragraph{R2R-CE}
Figure~\ref{fig:main-vis-r2r} illustrates instruction decomposition and landmark-grounded navigation. For the bathroom-mirror route, the agent ascends to the landing and uses the bedroom, bed, and closet as successive landmarks before approaching the mirror and verifying the stopping condition. For the patio-to-kitchen route, it first identifies the stair access, follows the planter to the steps, reaches the raised patio, and enters through the glass doors before continuing through the living room to the kitchen. The selected stages show how observation and panoramic inspection support the choice of grounded navigation targets and short-horizon instructions. Progress is assessed after each sub-goal, so the final stop is conditioned on reaching the requested endpoint. A navigation execution example and comparisons with InternVLA-N1 are provided in Figures~\ref{fig:vln-case} and~\ref{fig:r2r-ceExamples} in the appendix.

\begin{figure}[!htb]
\centering
\includegraphics[width=\textwidth]{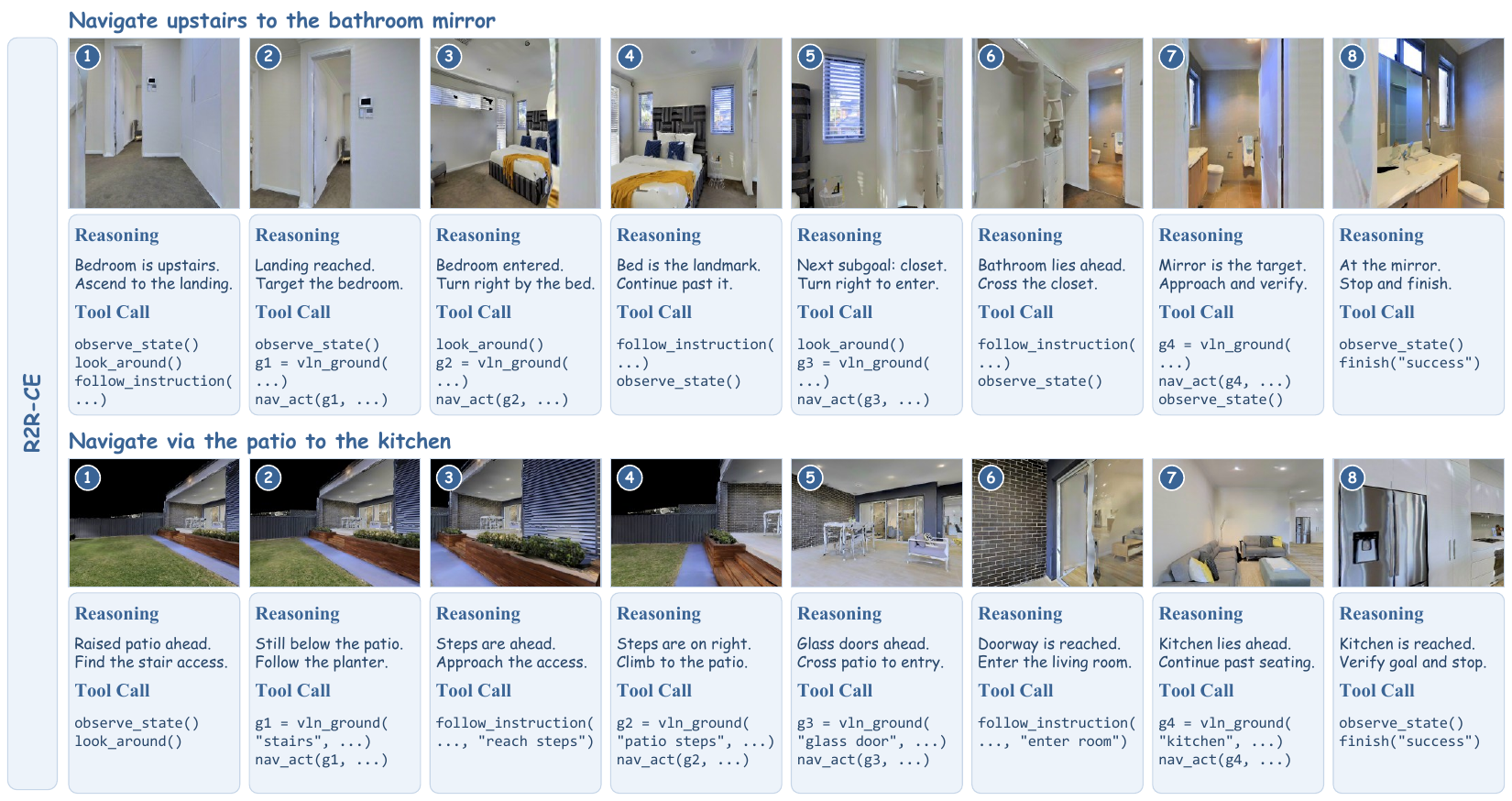}
\caption{\textbf{Examples of \method{} on R2R-CE Val-Unseen.} Top: navigating upstairs to a bathroom mirror. Bottom: navigating via a patio to a kitchen.}
\label{fig:main-vis-r2r}
\end{figure}

\par\noindent\begin{minipage}{\linewidth}
\paragraph{Real-world manipulation}
We evaluate transfer of the high-level planner to physical manipulation without additional fine-tuning on the real-world evaluation tasks. Figure~\ref{fig:real-world} presents two representative executions from the three-task suite evaluated in Figure~\ref{fig:realworld}. OmniBrain-0 coordinates perception, motion, and VLA-based skills in a closed loop, using pixel localization for block stacking and segmentation for apple placement. After each skill, the agent updates its execution state; transitions from \textit{held\_object=none} to \textit{held\_object=yellow cube} or \textit{apple} ground subsequent decisions in intermediate outcomes, supporting adaptive skill composition and feedback-driven completion.
\end{minipage}\par

\begin{figure}[!htb]
    \centering
    \includegraphics[width=1.0\textwidth]{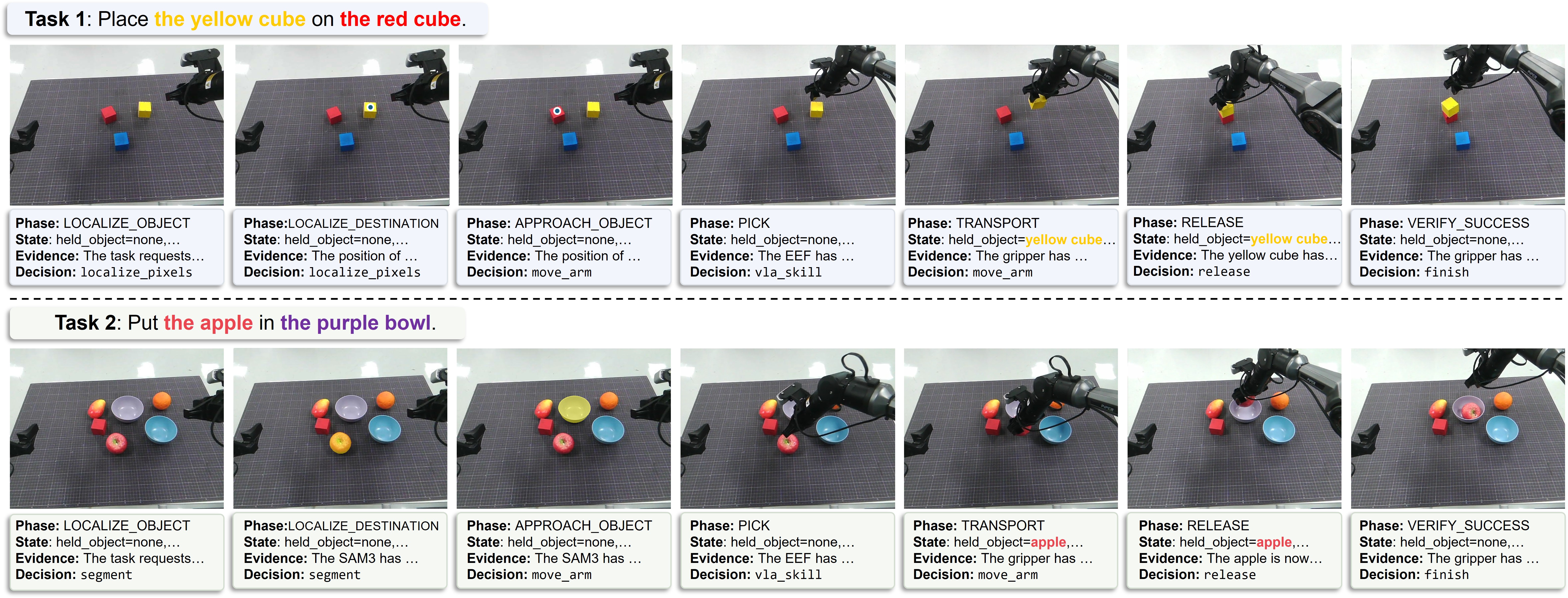}
    \caption{\textbf{Real-world manipulation examples.} Selected observations and structured decisions illustrate closed-loop perception, skill coordination, and execution feedback on a physical robot.}
    \label{fig:real-world}
    \end{figure}

\FloatBarrier
\subsubsection{Reminder-guided execution}
\label{sec:qualitative_reminder_execution}
Figures~\ref{fig:rsi_vis} and~\ref{fig:rsi_vis_vln} illustrate the inference-time reminder mechanism introduced in Section~\ref{sec:reminder_policy_iteration}. The reminder policy is checked against the current interaction context before the agent selects its next tool invocation. In the manipulation example (Figure~\ref{fig:rsi_vis}), the initial \texttt{move\_arm} and \texttt{vla\_skill} decisions proceed without reminders. Guidance is triggered at step~2 before \texttt{grasp} and again at step~3 before \texttt{move\_arm}, concentrating the intervention around grasping and the transition to subsequent motion. The later transport motions, release, and final \texttt{finish} decision proceed without further reminders. This pattern illustrates how the policy supplies guidance at selected states within an ongoing task, while \method{} continues to generate the tool calls. Aligning the observations, trigger states, and tool sequence makes the timing of reminder guidance explicit within closed-loop execution.

\begin{figure}[!htb]
    \centering
    \includegraphics[width=\textwidth]{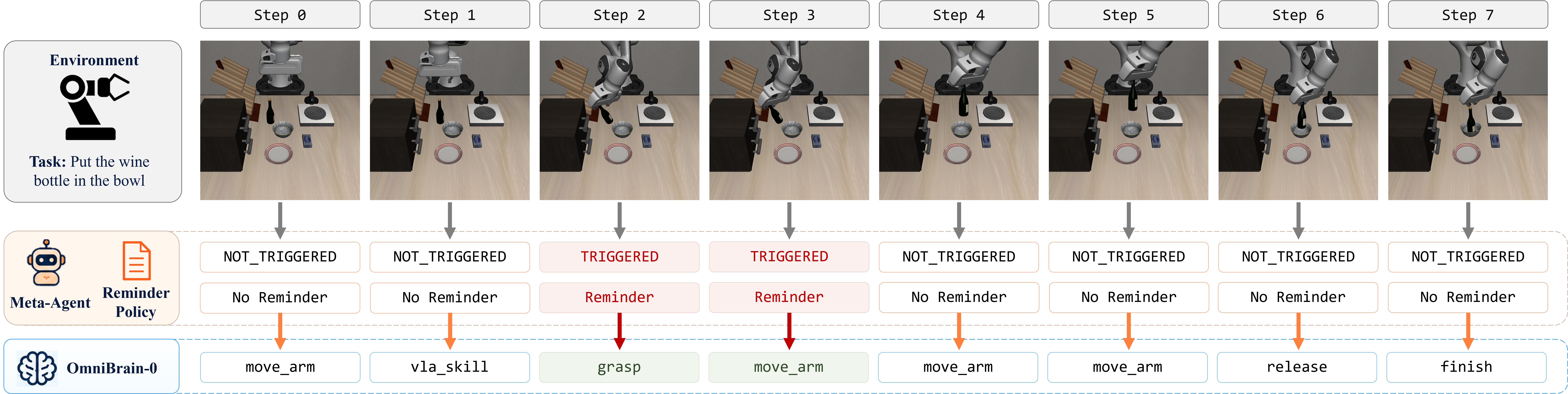}
    \caption{\textbf{Reminder-guided manipulation.} A wine-bottle placement example aligning visual observations (top), reminder-policy trigger states (middle), and \method{}'s tool invocations (bottom). For the task \textit{Put the wine bottle in the bowl}, reminders are triggered at steps~2 and~3 before \texttt{grasp} and \texttt{move\_arm}, respectively; the remaining displayed steps proceed without reminders. The sequence continues through bottle transport, release, and task termination.}
    \label{fig:rsi_vis}
\end{figure}

\begin{figure}[!htb]
    \centering
    \includegraphics[width=\textwidth]{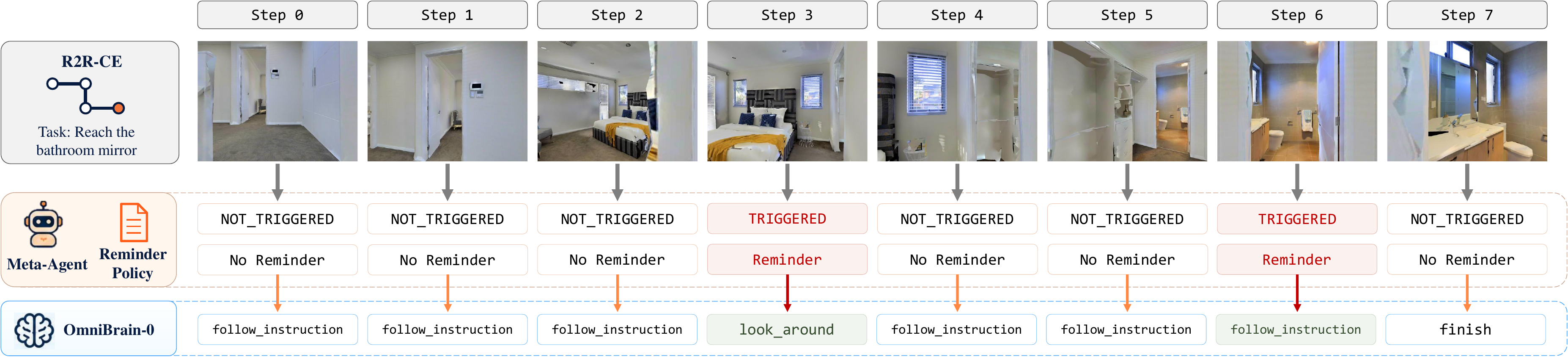}
    \caption{\textbf{Reminder-guided navigation.}
    An illustrative bathroom-mirror navigation example aligning visual observations (top), reminder-policy trigger states (middle), and \method{}'s tool invocations (bottom).
    Reminders at steps~3 and~6 guide route selection and prevent premature termination, respectively.}
    \label{fig:rsi_vis_vln}
\end{figure}

\FloatBarrier
\endgroup

\Needspace{18\baselineskip}
\section{Conclusion and Limitations}

We introduced \textbf{OmniBrain-0}, a compact unified embodied agent that coordinates manipulation and navigation through a shared high-level decision model. Our progressive training framework combines embodied pretraining, domain-specialized agent distillation, and multi-teacher on-policy distillation to transfer and consolidate planning, skill orchestration, and recovery capabilities. A structured data pipeline converts expert interactions into state-grounded supervision, retaining both canonical successes and successful recoveries. Rubric-Guided Recursive Self-Improvement further connects execution evidence, reminder-policy refinement, verified trajectory collection, and expert consolidation, supporting the continued construction of effective interaction data. Simulated manipulation and navigation benchmarks show strong task performance with substantially lower token consumption and shorter task completion times than the closed-source teacher, while spatial and real-world evaluations further assess grounding and physical applicability. These results support learning transferable embodied decision-making capabilities in a compact model while retaining specialized low-level skills.

OmniBrain-0 remains dependent on the coverage and reliability of its low-level skills, and gaps persist on unseen task compositions and spatial localization. Real-world evaluation also covers a limited set of tasks. Future work will broaden task and embodiment coverage, strengthen grounding and recovery under distribution shifts, and study sustained improvement through larger and more diverse verified interaction corpora.

\newcommand{\envelopeicon}{\ding{41}}

\clearpage
\section*{\hypertarget{contributions-heading}{Contributions and Acknowledgments}}
\phantomsection
\addcontentsline{toc}{section}{Contributions and Acknowledgments}
\label{sec:contributions}

\noindent
\begin{minipage}[t]{0.47\linewidth}
  \subsubsection*{Core Contributors}
  \begin{itemize}[leftmargin=*, itemsep=0.15em, topsep=0.25em]
    \item Yuechen Luo$^{*}$
    \item Chong He$^{*}$
    \item Fang Li$^{*},^{\dagger}$
    \item Shaoqing Xu$^{*},^{\dagger}$
    \item Jingkai Wang
    \item Menglun Su
    \item Jiaxin Liu
    \item Shengyin Jiang
    \item Zhiyi Lai
    \item Long Chen
    \item Hangjun Ye
    \item Fuxi Wen\textsuperscript{\envelopeicon}
    \item Zhi-xin Yang\textsuperscript{\envelopeicon}
  \end{itemize}
\end{minipage}%
\hfill
\begin{minipage}[t]{0.47\linewidth}
  \subsubsection*{Contributors}
  \begin{itemize}[leftmargin=*, itemsep=0.15em, topsep=0.25em]
    \item Longlong Wang
    \item Liang Hu
    \item Haochen Liu
    \item Yuncheng Jiang
    \item Hongwei Xie
    \item Jieming Cui
    \item Hanbing Li
    \item Junjie Huang
    \item Zhixiang Duan
    \item Hexin Zhang
    \item Weixiang Liang
    \item Ying Gong
    \item Yuhan Wang
    \item Yong Pan
    \item Pengna Li
    \item Kangyi Wu
    \item Ken Deng
    \item Kun Ma

  \end{itemize}
\end{minipage}

\vspace{0.5em}
\noindent{\small\itshape
$^{*}$Co-first authors (equal contribution).\quad
$^{\dagger}$Project Leaders.\quad
\textsuperscript{\envelopeicon}Corresponding authors.}\\
\vspace*{0.5em}
Technical details: \underline{luo-yc24@mails.tsinghua.edu.cn},~\underline{fang.lee@connect.um.edu.mo},~\underline{xushaoqing@xiaomi.com}

\subsubsection*{Acknowledgments}
\noindent
We thank the teams and maintainers of the models, datasets, benchmarks, simulation environments, and robotic platforms used in this work. These resources were used for research and evaluation purposes.

\clearpage
\bibliographystyle{plainnat}
\bibliography{main}

\clearpage
\beginappendix
\counterwithin{table}{section}
\counterwithin{figure}{section}

\noindent\textbf{Overview.}
This appendix provides additional details on the agentic system, evaluation, training data, and task execution.
\textbf{Section~\ref{appendix:agent_system}} presents the manipulation and navigation tools and the organization of the Experience Library.
\textbf{Section~\ref{appendix: eval}} specifies the benchmark settings, evaluation episodes, and metrics for simulated tasks, spatial understanding and referring, and real-world manipulation.
\textbf{Section~\ref{appendix:data_and_train}} describes the VQA and agentic data, including RSI-collected trajectories, together with expert collection, low-level model sizes, RSI model assignments, hardware, and training hyperparameters.
\textbf{Section~\ref{appendix:qualitative}} presents complete manipulation traces, a navigation execution example, and comparisons with InternVLA-N1.

\section{Details of the Agentic System Design} \label{appendix:agent_system}

This section provides additional details of the agentic system design, including the tools developed for the VLA and VLN tasks, as well as the construction and utilization of the Experience Library. The tools provide standardized interfaces for interacting with heterogeneous model-based and scripted skills, while the Experience Library organizes reusable knowledge and interaction experiences accumulated from expert models.

\paragraph{Tools}
We design a unified set of tools for VLA and VLN tasks, serving as the interface through which the agent interacts with and acts upon the physical environment. The detailed definitions of the tools are as follows.

\begin{figure}[!htbp]
    \centering
    \includegraphics[page=7,width=\textwidth]{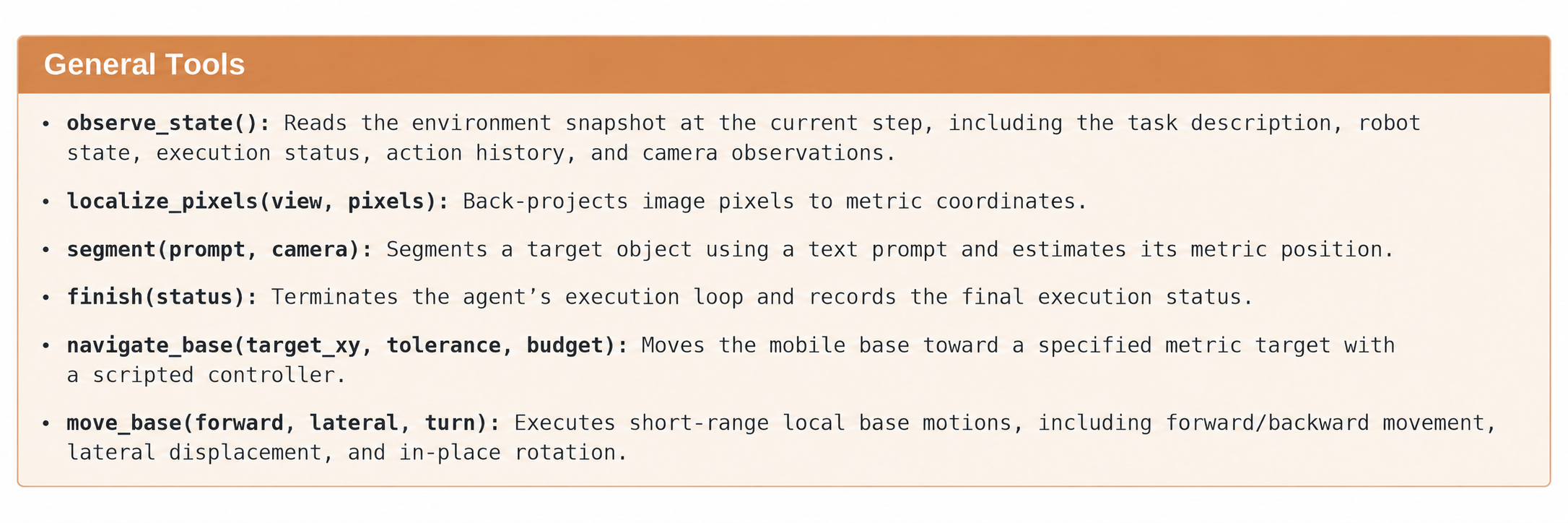}
    \caption{\textbf{General Tools:} Shared tools for environment
    observation, pixel-to-metric localization, text-guided object
    segmentation, execution termination, and mobile-base control.}
    \label{fig:general-tools}
\end{figure}

\begin{figure}[!htbp]
    \centering
    \includegraphics[page=8,width=\textwidth]{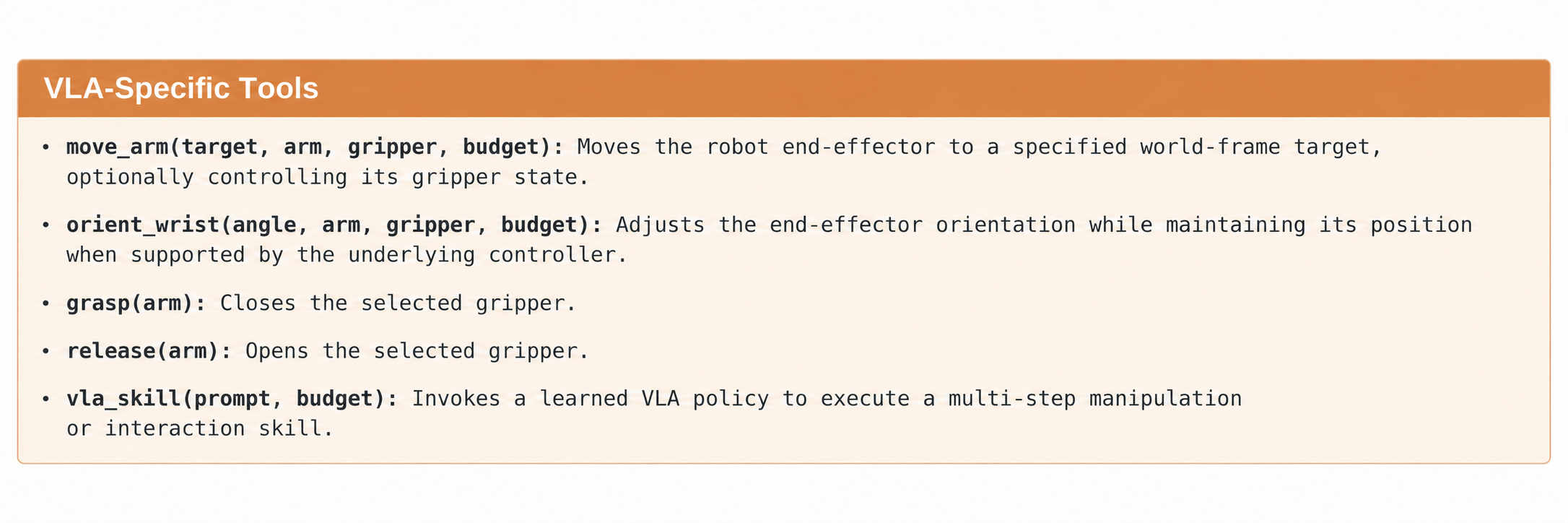}
    \caption{\textbf{VLA-Specific Tools:} Tools for end-effector motion,
    wrist orientation, gripper control, and multi-step manipulation
    or interaction through a learned VLA policy.}
    \label{fig:vla-tools}
\end{figure}

\begin{figure}[!htbp]
    \centering
    \includegraphics[page=9,width=\textwidth]{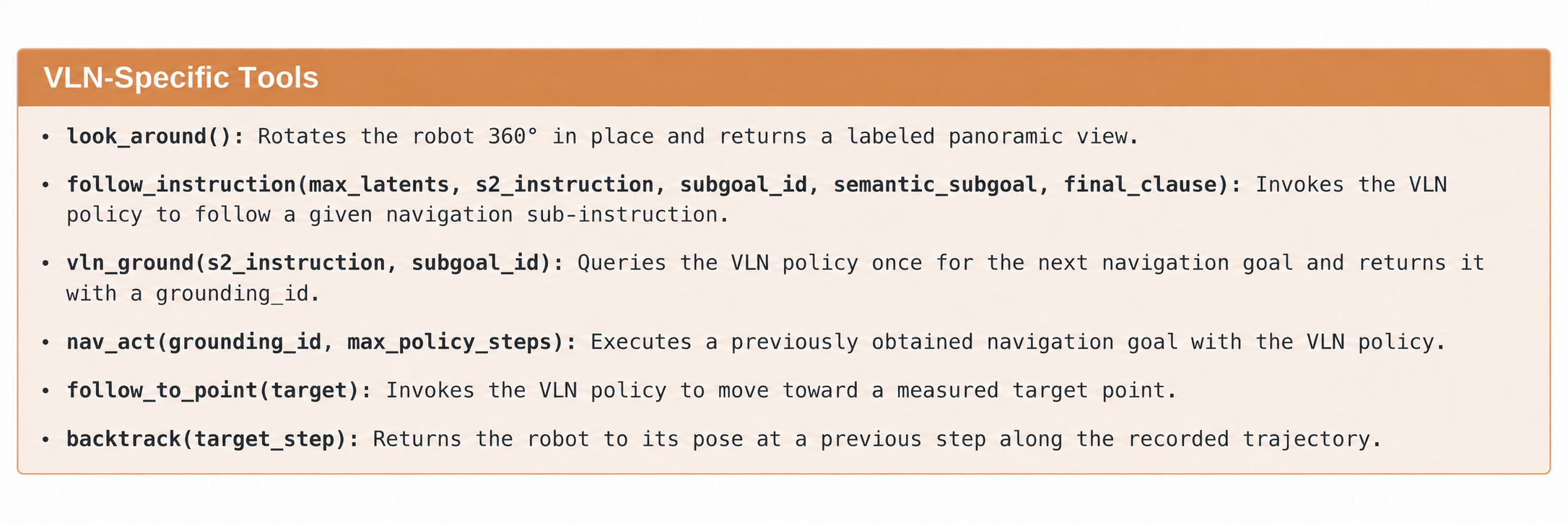}
    \caption{\textbf{VLN-Specific Tools:} Tools for panoramic observation,
    instruction following, navigation-goal grounding and execution,
    point-directed navigation, and trajectory backtracking.}
    \label{fig:vln-tools}
\end{figure}

\paragraph{Experience Library}
To collect diverse and successful expert interaction trajectories, we follow the data collection paradigm of Harness VLA~\cite{zhang2026harness}, allowing expert models to explore the same task environment across multiple episodes and autonomously reflect on the causes of failures and the factors contributing to successful executions. These cross-episode experiences are then manually reviewed and consolidated into an Experience Library, which provides reusable priors for subsequent model execution. The library is organized into four complementary categories, covering general principles, task execution patterns, failure recovery, and the capability boundaries of VLA/VLN skills.

\begin{enumerate}[leftmargin=*]
    \item \textbf{General Principles.}
    Reusable perception, coordinate, gripper, controller, navigation, and safety knowledge that applies across tasks. Examples include image-first decision making, object-to-end-effector offsets, appropriate camera/view selection, spatial landmark awareness, and short, capped contact or navigation motions.

    \item \textbf{Task Execution Patterns.}
    Reusable recipes for pick--transport--place, containment, articulated fixtures, stacking, handover, ranking, multi-object ordering, and navigation. These experiences describe task phases, skill composition, and verification gates rather than fixed scene coordinates, and cover both VLA-based manipulation and VLN-based navigation behaviors.

    \item \textbf{Failure Cases and Recovery Strategies.}
    Negative evidence and recovery strategies accumulated from unsuccessful executions, including common failure modes, their observable symptoms, recovery actions, and appropriate execution budgets. For VLA tasks, these may involve grasp failures, collisions, or inaccurate object localization; for VLN tasks, they include navigation deviations, incorrect waypoint selection, ambiguous landmark grounding, and ineffective recovery behaviors.

    \item \textbf{VLA/VLN Capability Boundaries.}
    Knowledge about the respective capabilities and limitations of VLA and VLN skills, including the conditions under which to invoke learned VLA/VLN policies, switch between them, or fall back to scripted tools. These experiences capture when learned skills are more reliable and when explicit geometric or rule-based actions are preferable.
\end{enumerate}

\section{Benchmarks and Evaluation Protocol} \label{appendix: eval}

We use a common episode-level protocol across the closed-loop benchmarks. Each episode is associated with its benchmark, task suite, task, perturbation, random seed, and model version. Duplicate records of the same evaluation case are removed before aggregation. The planner receives the task instruction, camera observations, robot state or proprioception, and tool feedback. Hidden task specifications, ground-truth object poses, and evaluation labels are used for scene construction or scoring and are not supplied to the planner. The planner also has no access to the environment state of other evaluation episodes.

An episode ends when the environment confirms success, the planner requests termination, an execution limit is reached, or an infrastructure failure prevents further interaction. Completion claimed by the planner is accepted only when supported by the environment's success criterion. Unless a benchmark specifies a macro-average, Overall is computed as the total number of successful episodes divided by the total number of evaluated episodes. Task failures, timeouts, simulator initialization failures, infrastructure errors, and invalid or missing results are recorded separately. When infrastructure failures are retried, both the original attempt and the retry are retained in the evaluation records.

\paragraph{LIBERO-Pro}
LIBERO-Pro~\cite{zhou2025liberopro} is an extension of the LIBERO benchmark designed to evaluate the generalization capabilities of vision-language-action (VLA) models. It introduces controlled perturbations to the original LIBERO tasks, including changes to manipulated objects, object positions, task instructions, task configurations, and environments. Following Harness VLA~\cite{zhang2026harness}, we evaluate eight LIBERO-Pro conditions: \textsc{Spatial-T}, \textsc{Spatial-S}, \textsc{Object-T}, \textsc{Object-S}, \textsc{Goal-T}, \textsc{Goal-S}, \textsc{L10-T}, and \textsc{L10-S}, where \textsc{T} denotes task or instruction redirection and \textsc{S} denotes spatial swapping of object configurations. Each condition contains ten tasks evaluated with seeds $s_1$--$s_{10}$, yielding 100 episodes per condition and 800 episodes overall.

\paragraph{LIBERO-Plus}
LIBERO-Plus~\cite{fei2025liberoplus} extends LIBERO with 10,030 task variants spanning camera viewpoints, robot initial states, language instructions, lighting, background textures, sensor noise, and object layout. These variations assess robustness to visual, linguistic, and physical changes in manipulation settings. For LIBERO-Plus, we evaluate four suites: \textsc{Spatial}, \textsc{Object}, \textsc{Goal}, and \textsc{LIBERO-10}. From each suite, we sample 210 task variants, evenly distributed across seven perturbation types: \textsc{Camera}, \textsc{Robot}, \textsc{Language}, \textsc{Light}, \textsc{Background}, \textsc{Noise}, and \textsc{Layout}. Each perturbation type contains 30 tasks per suite, all evaluated with seed $s_0$, yielding 840 episodes overall.

\paragraph{RoboTwin 2.0}
RoboTwin 2.0~\cite{chen2025robotwin} is a simulation benchmark for evaluating generalist policies in bimanual robotic manipulation, featuring 50 diverse manipulation tasks. It provides two evaluation settings: a \texttt{Clean} setting (Easy), which largely preserves the standard environment, and a \texttt{Randomized} setting (Hard), which introduces substantial environmental variations. This design enables evaluation of both task execution capability and robustness to distribution shifts. Following Harness VLA~\cite{zhang2026harness}, we evaluate all 50 tasks under both the \texttt{Clean} and \texttt{Randomized} settings, with five seeds per task in each setting, resulting in $2\times50\times5=500$ episodes. The planner receives task instructions, camera observations, robot state, and access to the available bimanual tools and skills. Scene randomization and ground-truth simulator state are used to construct and evaluate episodes; hidden evaluation labels are not provided to the planner. Success is determined by the simulator's built-in task completion criterion. We report separate success rates for Clean and Randomized environments and an Overall rate across all 500 episodes.

\paragraph{RoboCasa365}
RoboCasa365~\cite{nasiriany2026robocasa365} is a large-scale simulation benchmark for evaluating generalist robot policies on everyday household tasks. The benchmark includes both atomic tasks targeting individual manipulation skills and composite tasks that require sequencing multiple skills for longer-horizon activities, enabling evaluation of general task-solving capabilities and generalization across diverse environments and task compositions. Following Harness VLA~\cite{zhang2026harness}, we evaluate on the RoboCasa365 Target-50 split, comprising 18 \textsc{Atomic-Seen}, 16 \textsc{Composite-Seen}, and 16 \textsc{Composite-Unseen} tasks. The \textsc{Seen} groups correspond to task templates present in pretraining, while \textsc{Unseen} contains templates held out from pretraining. We use held-out seeds $s_1$--$s_{10}$ for \textsc{Atomic-Seen} and $s_1$--$s_5$ for both composite groups, yielding $18\times10=180$, $16\times5=80$, and $16\times5=80$ episodes, respectively, for 340 episodes overall. RoboCasa365 uses a task-macro Overall: each task first receives a success rate computed over its evaluation episodes, and the 50 task rates are then averaged with equal weight. Thus, the three split groups contribute 18, 16, and 16 task rates, respectively, and this Overall differs from the episode-micro average used by the general rule above.

\paragraph{R2R-CE}
R2R-CE~\cite{krantz2020beyond} is the continuous-environment adaptation of the Room-to-Room (R2R) vision-and-language navigation benchmark. It ports language-annotated navigation episodes to the Habitat simulator, where an agent follows a natural-language route instruction through reconstructed indoor environments. Unlike the original navigation-graph setting, the agent moves through continuous space rather than jumping between predefined viewpoints. We evaluate a fixed subset of 100 episodes from the Val-Unseen split. The planner receives the navigation instruction, camera observations, and execution feedback, and uses SmartWay~\cite{shi2025smartway} for navigation. Ground-truth shortest paths and goal positions are used for scoring only. We report Navigation Error (NE), the final distance to the goal; Success Rate (SR), the percentage of episodes ending within 3\,m of the goal; Oracle Success Rate (OSR), the percentage whose trajectory reaches within 3\,m of the goal at any point; and Success weighted by Path Length (SPL), which accounts for both success and path efficiency. All four metrics are computed on the same 100 episodes, with service failures, invalid actions, timeouts, and navigation failures recorded separately.

\paragraph{EmbSpatial-Bench}
EmbSpatial-Bench~\cite{du2024embspatial} contains 3640 multiple-choice visual questions derived from MP3D, AI2-THOR, and ScanNet scenes. It tests six egocentric relations: above, below, left, right, close, and far. MP3D contributes 1201 questions, AI2-THOR contributes 1239, and ScanNet contributes 1200. Each input consists of an image, a spatial question, and candidate answers. Accuracy is the percentage of questions answered correctly. Table~\ref{tab:embspatial} reports accuracy for each source environment and across all questions; Overall therefore weights each environment by its question count.

\paragraph{RefSpatial-Bench}
RefSpatial-Bench~\cite{zhou2026roborefer} evaluates spatial referring from an image and a spatially constrained instruction. Its Location and Placement sets contain 100 examples each, requiring target-object localization and free-space selection, respectively. Unseen comprises 77 examples drawn from these sets with spatial-relation combinations absent from the benchmark authors' RefSpatial training data. A prediction is correct when the output point falls within the annotated target mask. Table~\ref{tab:refspatial} reports the percentage of correct predictions for Location, Placement, and Unseen. Overall weights each category by its size: 100 examples for Location, 100 for Placement, and 77 for Unseen. Because Unseen overlaps the two task sets, Overall summarizes the evaluation categories rather than accuracy on a disjoint set of 277 examples.

\paragraph{Real-world manipulation}
The real-world manipulation suite comprises three language-conditioned tasks: \textit{block stacking}, where the robot places the yellow cube on the red cube; \textit{apple placement}, where the robot picks up the apple and places it in the purple bowl; and \textit{drawer opening}, where the robot grasps the handle and pulls the drawer open. These tasks cover precise object placement, pick-and-place manipulation, and contact-rich articulated-object manipulation. Each method is evaluated over 20 independent trials per task. A trial is successful if the robot completes the task within the allotted execution budget: placing the red cube stably on the blue cube, placing and releasing the apple inside the blue bowl, or opening the drawer to the specified target position. We report per-task Success Rate (SR), defined as the percentage of successful trials, and the macro-average SR across the three tasks.

\section{Datasets and Experimental Details} \label{appendix:data_and_train}

\paragraph{VQA Data} 
We construct our embodied pretraining corpus from EmbSpatial-SFT~\cite{du2024embspatial}, RefSpatial~\cite{zhou2026roborefer}, ShareRobot~\cite{ji2025robobrain}, and RoboInter-VQA~\cite{li2026robointer}. EmbSpatial-SFT provides instruction-tuning data for egocentric spatial relationship understanding and object localization, covering six fundamental spatial relations. RefSpatial further enriches spatial understanding and referring with 2.5M samples and 20M question-answer pairs from 2D images, 3D embodied videos, and simulated scenes, spanning 31 spatial relations and multi-step spatial reasoning. ShareRobot provides heterogeneous supervision for task planning, object affordance, and end-effector trajectories, spanning 23 source datasets, 12 embodiments, and 107 atomic task types, with over 1.0M planning-oriented VQA pairs. RoboInter-VQA complements these datasets with approximately 2.3M embodied VQA samples covering spatial and temporal reasoning, intermediate-representation understanding, and task planning. From these sources, we select samples focusing on spatial grounding, spatial reasoning, affordance understanding, and task planning, and incorporate agentic interaction data at a predefined ratio, ultimately obtaining approximately 820K embodied pretraining samples.

\paragraph{Agentic Data}
The interaction records preserve task instructions, visual observations, robot states, decision text, tool calls, and environment feedback in temporal order. We apply the data pipeline in Section~\ref{sec:data_pipeline} to verify task completion, remove incomplete, invalid, or noisy records, and normalize retained trajectories into multi-turn multimodal conversations. Each decision and tool invocation is aligned with the observations and feedback available at that step. The resulting corpus combines successful trajectories from the initial expert collection with additional verified successful trajectories collected through the RSI Data Flywheel. LIBERO/LIBERO-Pro, RoboTwin~2.0, RoboCasa365, and R2R-CE contribute 2,349, 513, 880, and 1,246 usable expert demonstrations, respectively. Supervision targets the planner's decision text and tool invocations, with observations and environment feedback as conditioning context.

\begin{figure}[!htbp]
    \centering
    \includegraphics[width=\textwidth]{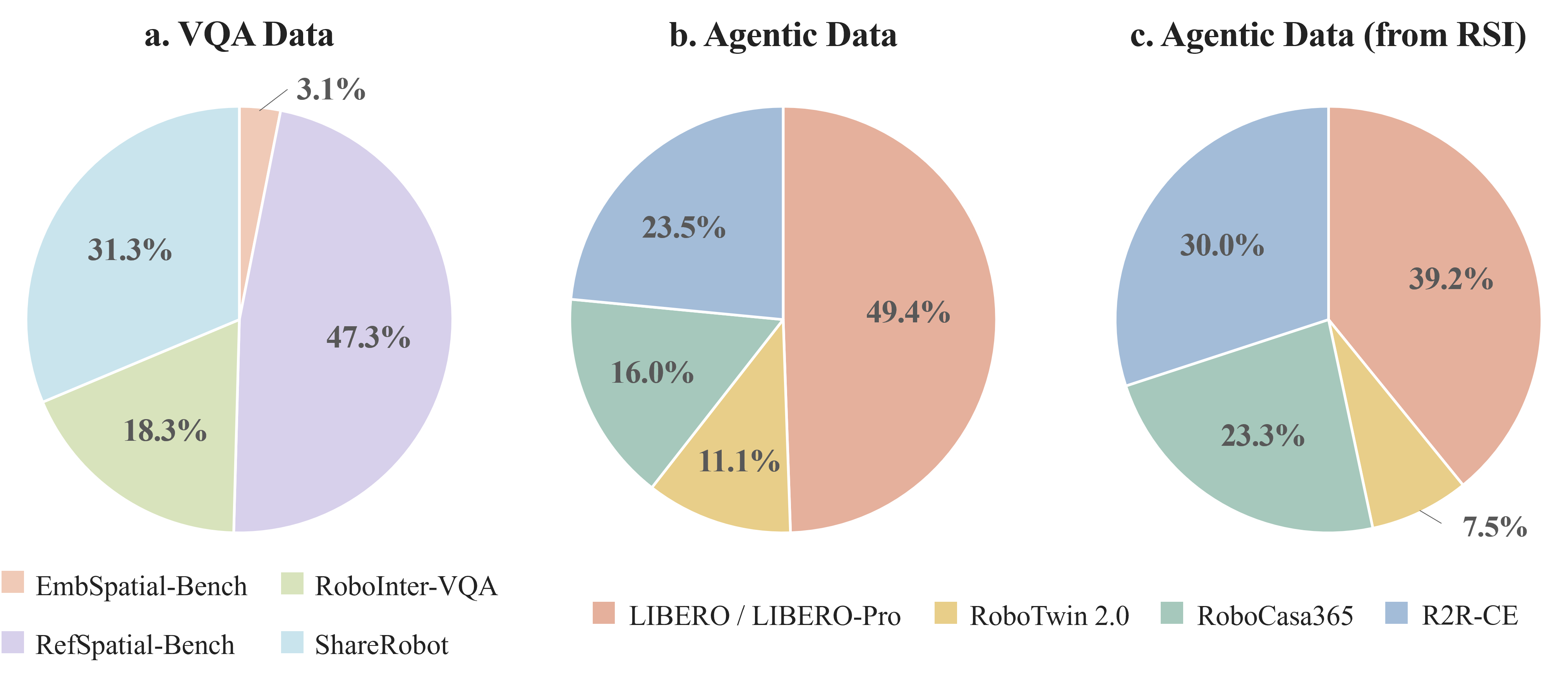}
    \caption{\textbf{Training data composition.} (a) The VQA corpus contains 820,453 samples. (b) The initial agentic corpus contains 3,846 demonstrations. (c) RSI contributes 1,142 additional demonstrations. Percentages indicate each source's share within the corresponding corpus.}
    \label{fig:training_data_composition}
\end{figure}

Figure~\ref{fig:training_data_composition} summarizes the source composition of the VQA corpus and the initial and RSI-collected agentic corpora.

\paragraph{Expert Collection and Execution Configuration}
We use GPT-5.6-Sol and GLM-5.3-Flash to collect closed-loop interaction trajectories on LIBERO/LIBERO-Pro, RoboTwin~2.0, RoboCasa365, and R2R-CE. At each decision step, an expert receives visual observations and available state feedback, together with an experience library that provides reusable guidance on task organization, tool composition, and failure recovery. It selects a tool and its parameters, then conditions its next decision on the resulting observations and execution status. Following the tool-based execution paradigm of Harness VLA~\cite{zhang2026harness}, a shared invocation-and-feedback interface separates high-level decisions from benchmark-specific low-level execution. Available manipulation skills include $\pi_{0.5}$~\cite{pmlr-v305-black25a}, LingBot-VLA~\cite{wu2026pragmatic}, and RLDX-1~\cite{rldx2026}; navigation uses SmartWay~\cite{shi2025smartway}. Detailed tool definitions are given in Appendix~\ref{appendix:agent_system}.

\paragraph{Low-Level Model Sizes}
The manipulation policies used in our experiments contain approximately 3.617B parameters for $\pi_{0.5}$, 4.197B for LingBot-VLA, and 6.913B for RLDX-1. These counts refer to the complete deployed policy weights, including vision and action components, rather than only the language backbone. For navigation, SmartWay's waypoint predictor contains 54M parameters; this count applies to the waypoint prediction component alone. InternVLA-N1, used as a navigation comparison model in the qualitative experiments, has 7B parameters. Here, B and M denote billions and millions of parameters, respectively.

\paragraph{RSI Model Configuration}
GPT-6.1-Sol serves as the Meta-Agent for rubric analysis and reminder-policy refinement in Section~\ref{sec:reminder_policy_iteration}. GPT-5.6-Sol serves as the judge model for reminder-policy evaluation.

\paragraph{Hardware Configuration and Training Parameters} We train Qwen3.5-4B/9B on 16 NVIDIA H20 GPUs using bfloat16 precision. Embodied Pretraining runs for $2$ epochs with a batch size of $1$, a learning rate of $5 \times 10^{-6}$, $4$ gradient accumulation steps, and utilizes the Adam optimizer with a weight decay of $0.05$ and a warmup ratio of $0.1$. We then perform Domain-Specialized Distillation for $3$ epochs using a batch size of $2$, a learning rate of $1 \times 10^{-5}$, $2$ gradient accumulation steps, and the same Adam optimizer setup (weight decay $0.05$, warmup ratio $0.1$). Finally, MOPD is applied for $2$ epochs with a batch size of $2$, a learning rate of $2 \times 10^{-6}$, and $2$ gradient accumulation steps. Multiple frozen teacher models are independently deployed on dedicated GPUs. MOPD-specific configurations include a rollout temperature of $1.0$, a distillation support top-$k$ of $64$, a Reverse KL distillation objective, and $\rho_{VLA}=\rho_{VLN}=0.5$.

\section{More Qualitative Examples}
\label{appendix:qualitative}

\begin{figure}[!htbp]
    \centering
    \includegraphics[width=\textwidth]{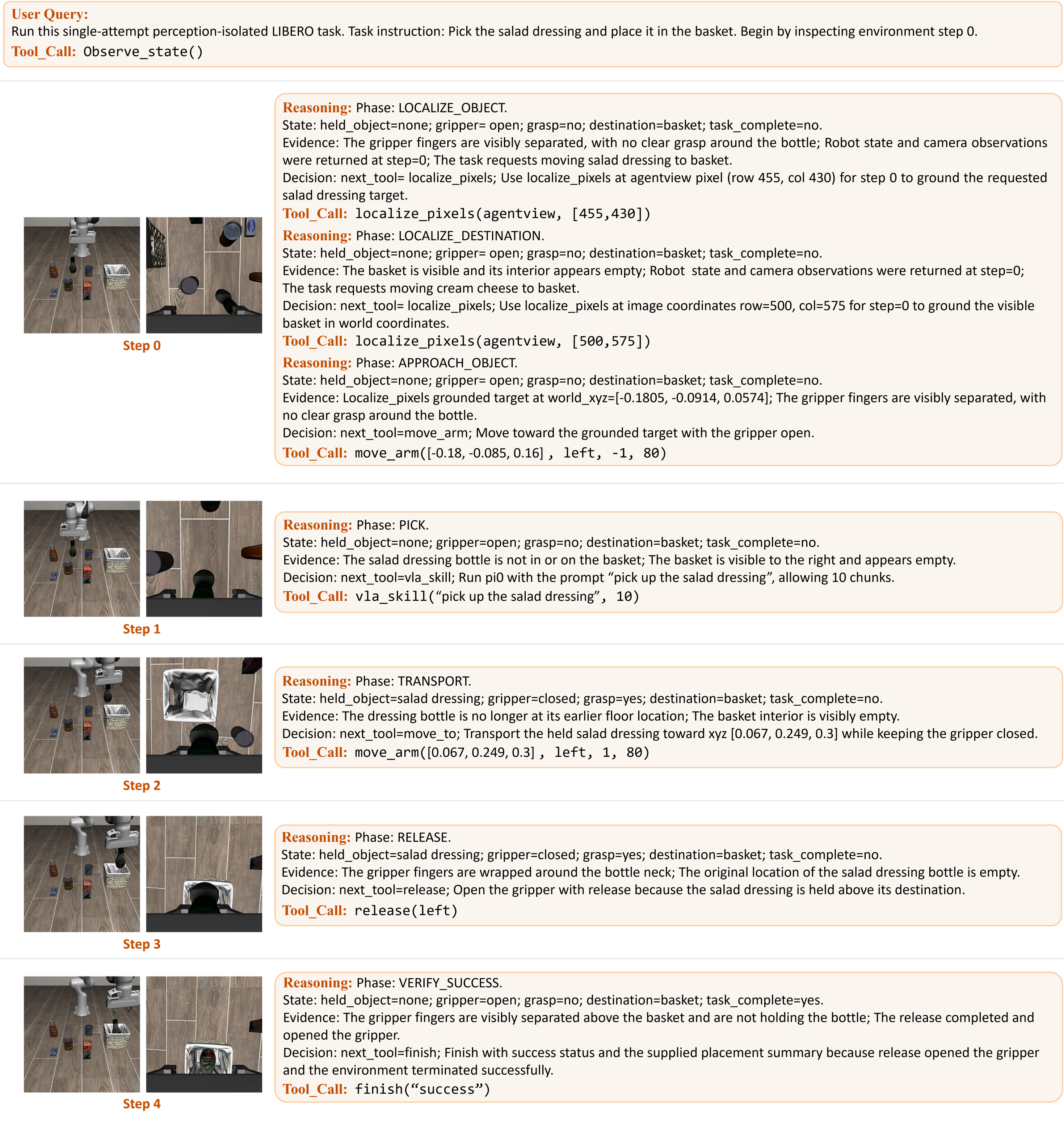}
    \caption{\textbf{LIBERO-Pro task execution.} Complete execution process for the task \textit{Pick the salad dressing and place it in the basket}.}
    \label{fig:vis_task_libero}
\end{figure}

\begin{figure}[!htbp]
    \centering
    \includegraphics[width=\textwidth]{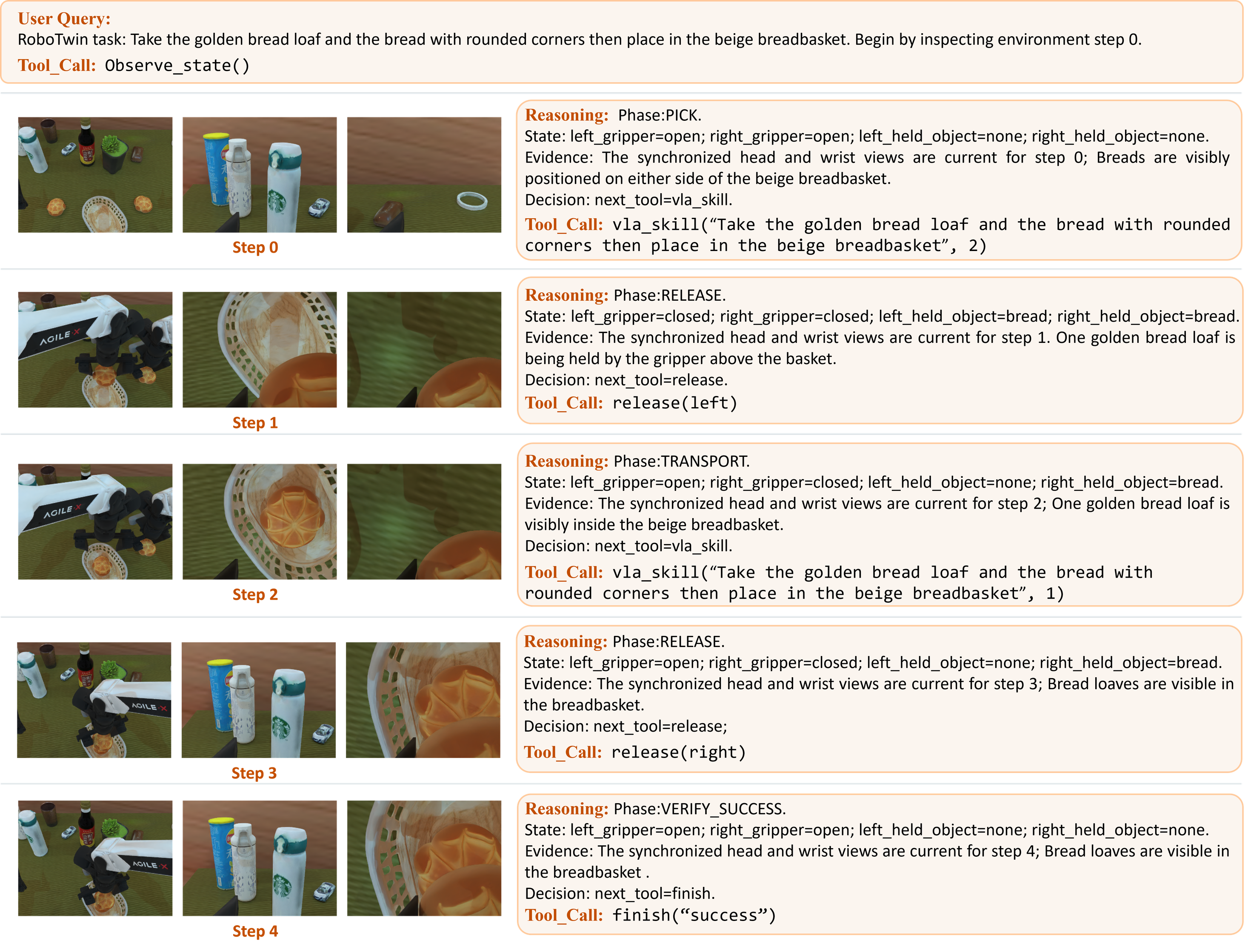}
    \caption{\textbf{RoboTwin 2.0 task execution.} Complete execution process for placing the two specified bread loaves in the beige breadbasket.}
    \label{fig:vis_task_robotwin}
\end{figure}

\begin{figure}[!htbp]
    \centering
    \includegraphics[height=0.82\textheight,keepaspectratio]{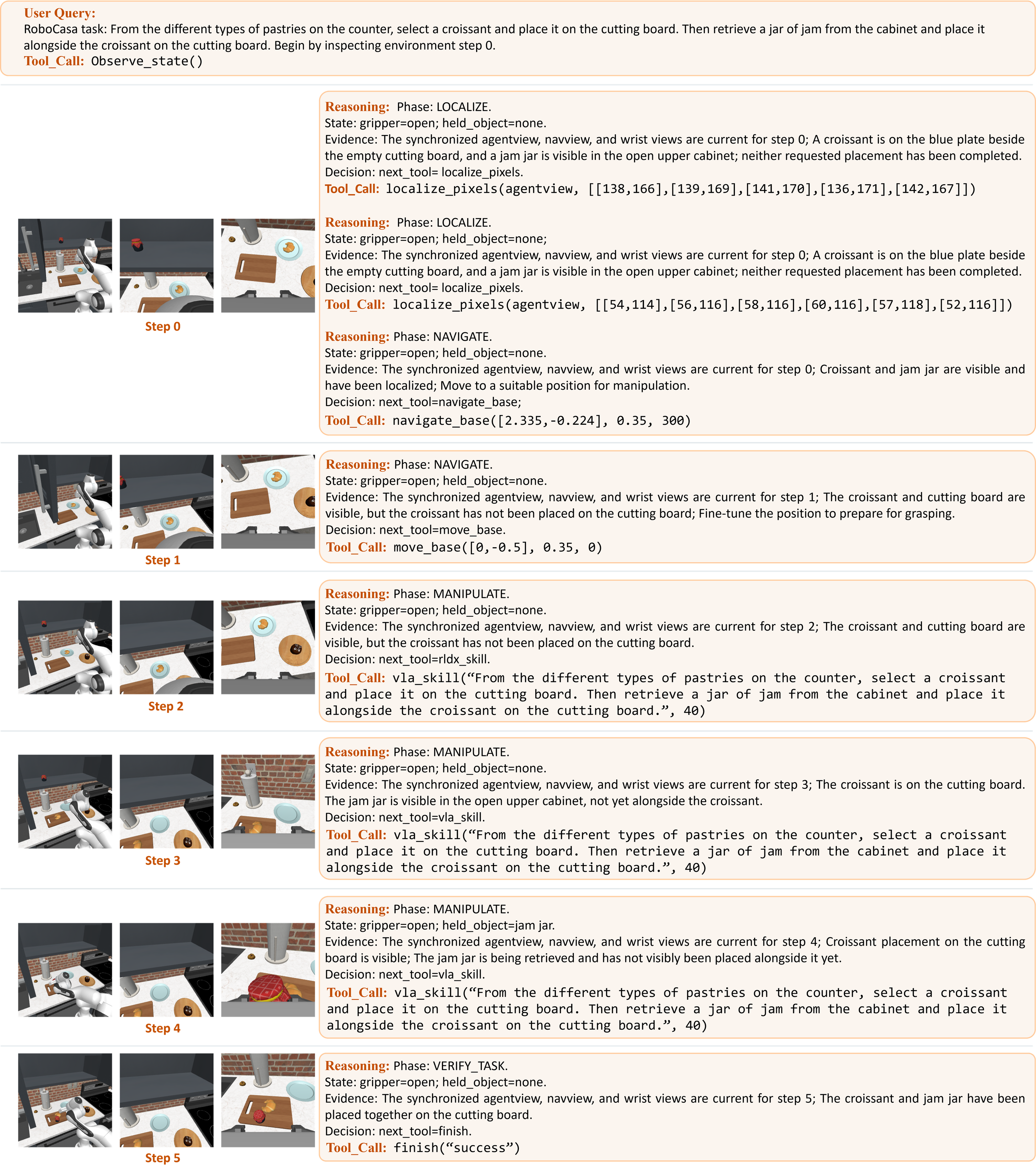}
    \caption{\textbf{RoboCasa365 task execution.} Complete execution process for placing a croissant and a jam jar together on the cutting board.}
    \label{fig:vis_task_robocasa}
\end{figure}

\begin{figure}[!htbp]
    \centering
    \includegraphics[width=\textwidth,height=0.78\textheight,keepaspectratio]{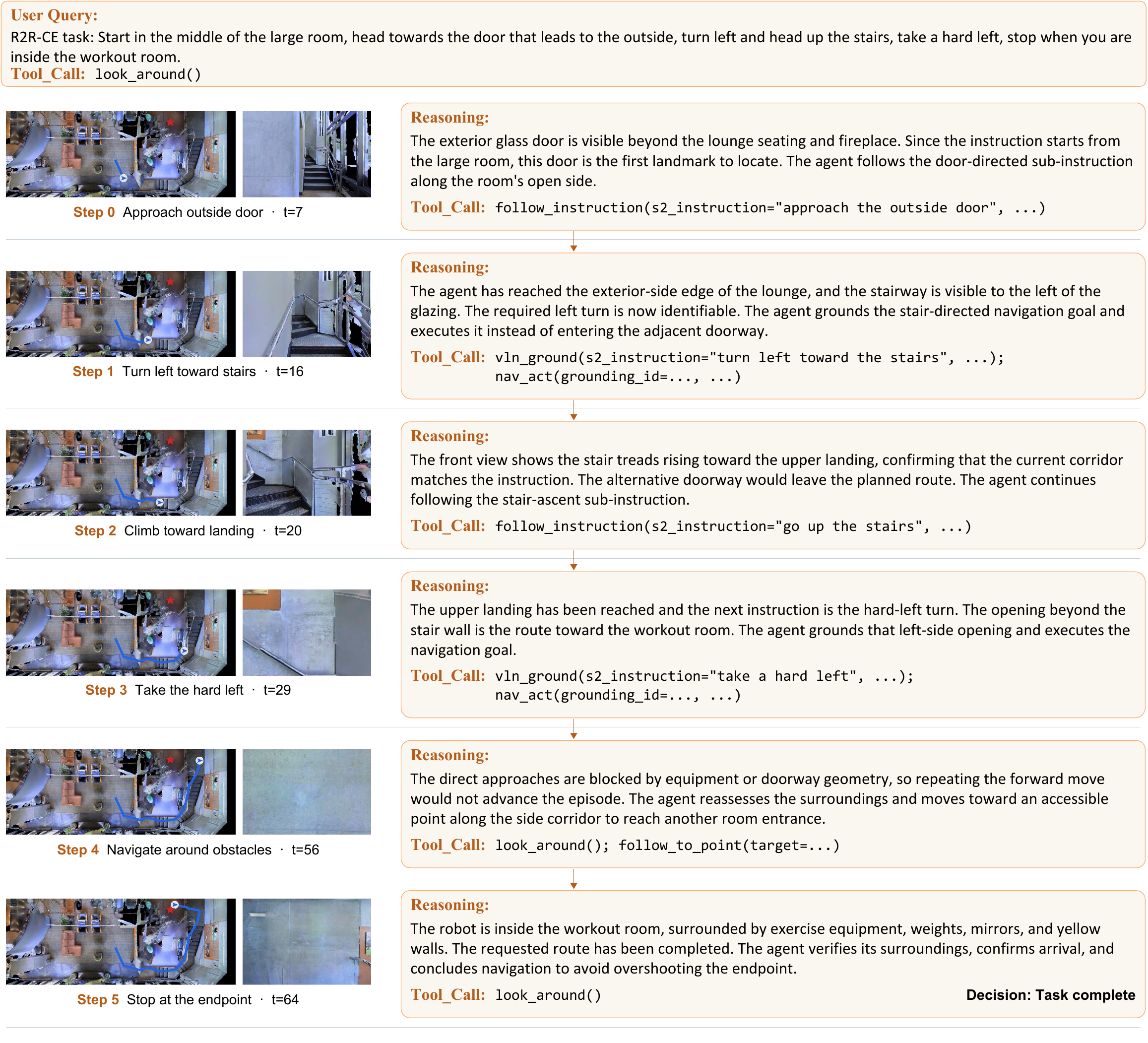}
    \caption{\textbf{R2R-CE task execution.} Selected stages of a navigation task in the Val-Unseen split, illustrating landmark grounding, instruction following, obstacle avoidance, and endpoint verification.}
    \label{fig:vln-case}
\end{figure}

\begin{figure}[!htbp]
    \centering
    \includegraphics[width=\textwidth,height=0.26\textheight,keepaspectratio]{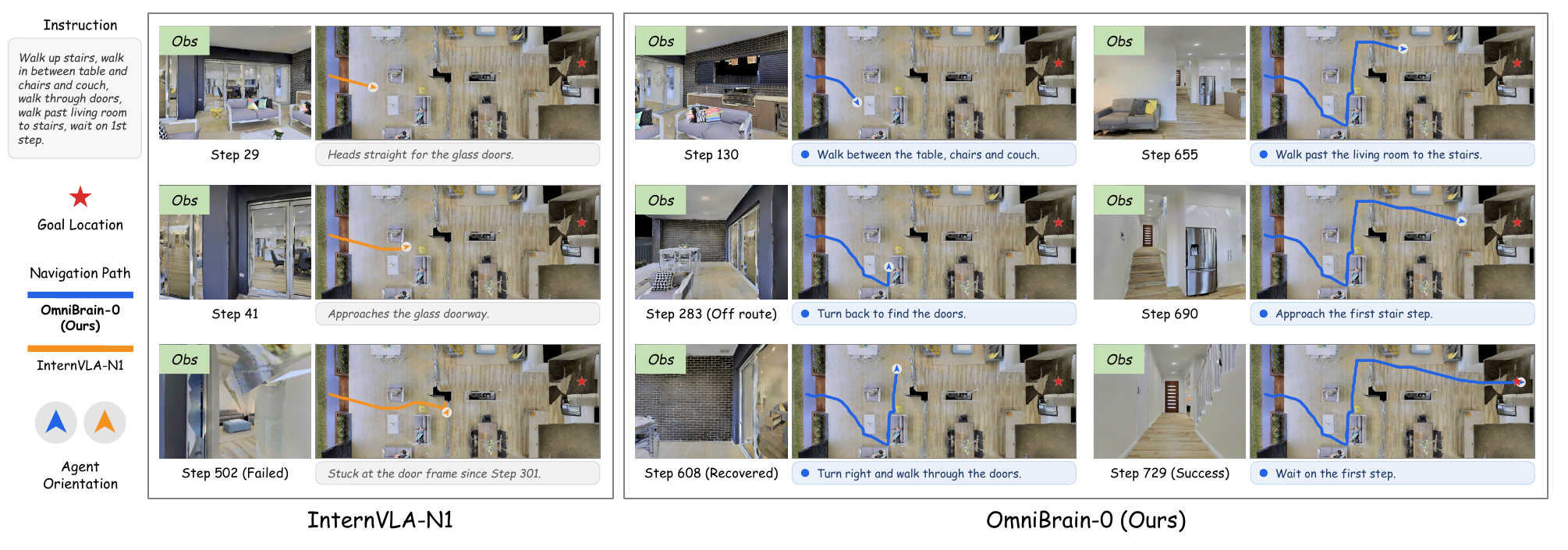}
    \includegraphics[width=\textwidth,height=0.26\textheight,keepaspectratio]{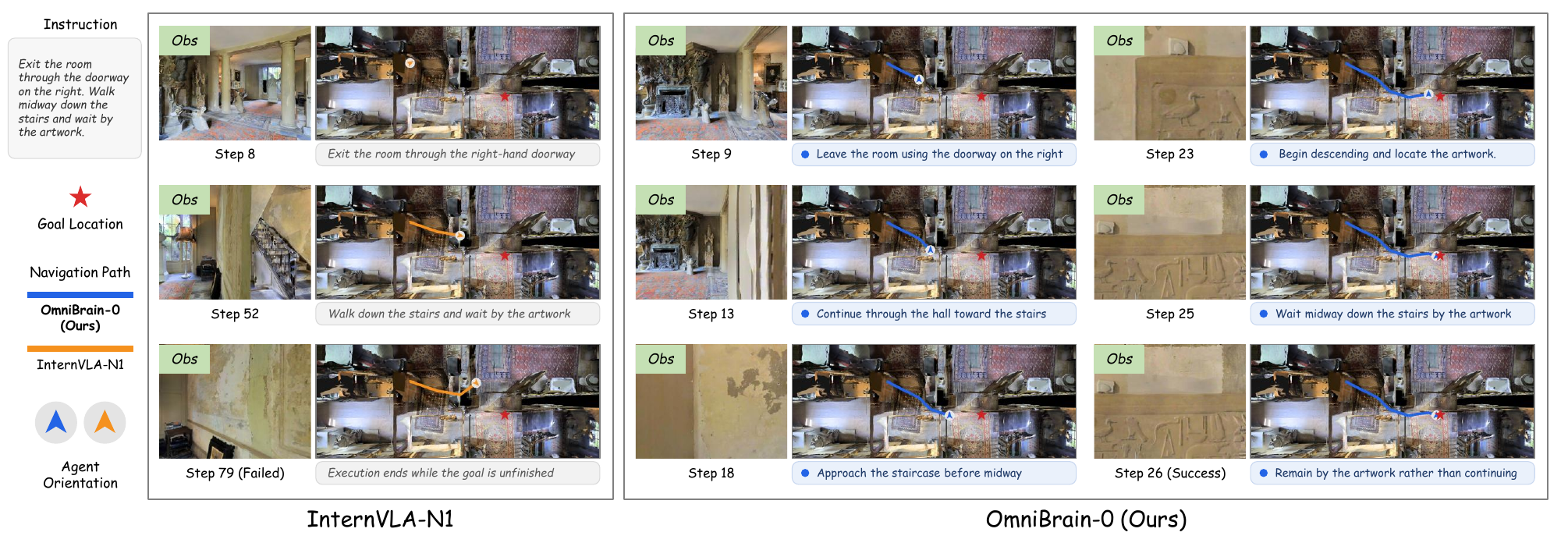}
    \includegraphics[width=\textwidth,height=0.26\textheight,keepaspectratio]{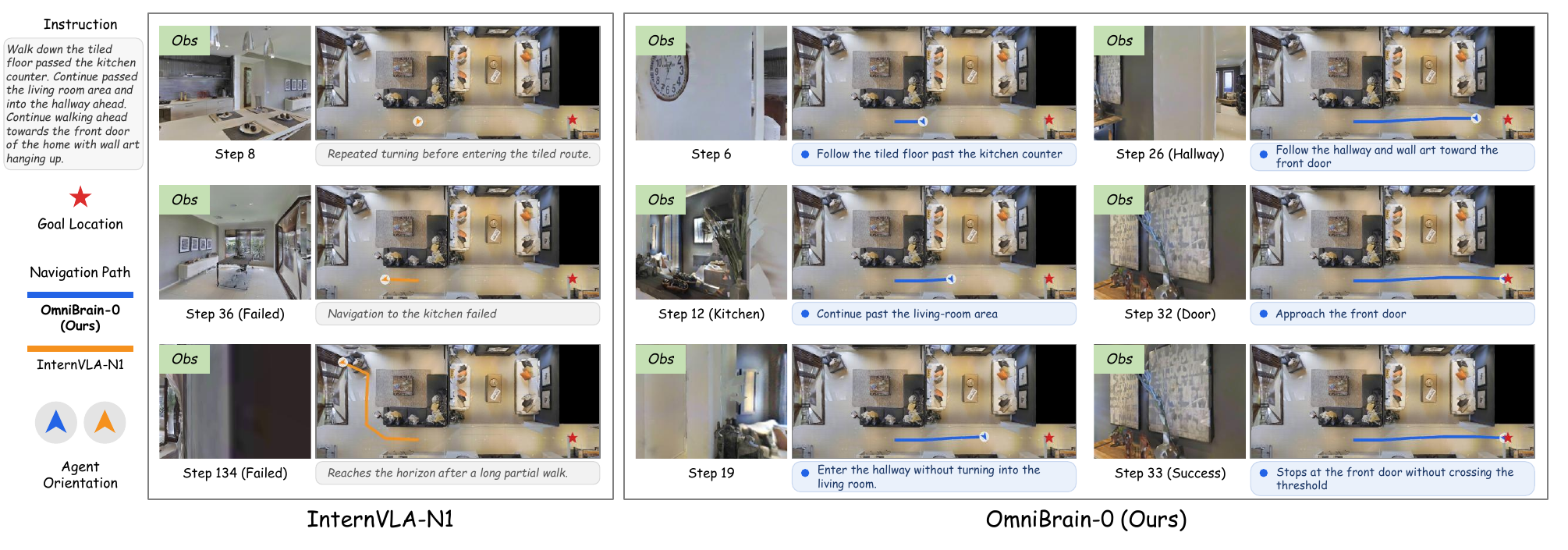}
    \caption{\textbf{R2R-CE navigation comparisons.} Each sequence compares InternVLA-N1 (left) and \method{} (right) on the same Val-Unseen task. Cells pair front-view observations with top-down trajectories at the indicated environment steps. Blue annotations show \method{}'s decisions, and gray annotations describe InternVLA-N1's behavior.}
    \label{fig:r2r-ceExamples}
    \end{figure}

This section complements the benchmark-level visualizations in Section~\ref{sec:qualitative_analysis} with complete manipulation execution traces and detailed navigation comparisons. Figures~\ref{fig:vis_task_libero}--\ref{fig:vis_task_robocasa} show the observation--reasoning--tool interaction sequences for three manipulation tasks. Figure~\ref{fig:vln-case} presents selected stages of R2R-CE task execution, and Figure~\ref{fig:r2r-ceExamples} compares navigation with InternVLA-N1. The navigation keyframes illustrate route decisions and outcomes rather than every intermediate action.

\paragraph{Task execution process of \method{}}
Figures~\ref{fig:vis_task_libero}, \ref{fig:vis_task_robotwin}, and~\ref{fig:vis_task_robocasa} present complete task execution processes across LIBERO-Pro, RoboTwin~2.0, and RoboCasa365; Figure~\ref{fig:vln-case} extends the analysis to R2R-CE task execution. At each manipulation turn, the agent produces structured reasoning with four fields, \textit{Phase}, \textit{State}, \textit{Evidence}, and \textit{Decision}, and then issues a tool call. These fields connect the current task stage and interaction state to the evidence supporting the next action. The traces show object and destination grounding, specialist-skill invocation, intermediate state tracking, and verified termination. The navigation example likewise connects observations and route progress to tool selection and endpoint verification.

\paragraph{LIBERO-Pro: grounded pick-and-place}
In Figure~\ref{fig:vis_task_libero}, the agent localizes the salad dressing and the basket separately, then approaches the target with the gripper open. A VLA skill establishes the grasp, after which explicit arm motion transports the held bottle above the basket. The state changes from holding the dressing to an open, empty gripper after release. The final decision uses the placement outcome and environment success signal to terminate, illustrating how localization, grasping, transport, and verification form a closed-loop sequence.

\paragraph{RoboTwin 2.0: tracking two held objects}
Figure~\ref{fig:vis_task_robotwin} requires placing two specified bread loaves in the beige breadbasket. The agent tracks the left and right grippers separately. After the initial VLA execution, both grippers hold bread; releasing the left loaf completes only part of the task. The next decision retains the right-hand grasp, invokes the manipulation skill to continue placement, and then releases the right loaf. Termination follows visual confirmation that both loaves are in the basket and both grippers are empty.

\paragraph{RoboCasa365: preserving multi-stage task progress}
In Figure~\ref{fig:vis_task_robocasa}, the agent must place a croissant on the cutting board and then retrieve a jam jar from the cabinet to place beside it. It first grounds the objects and adjusts the mobile base for manipulation. Subsequent decisions distinguish the completed croissant placement from the remaining jam-jar sub-goal, and track the jar while it is held and transported. The final check considers both objects together on the board, illustrating progress tracking across navigation, manipulation, and successive placement requirements.

\paragraph{R2R-CE: task execution}
Figure~\ref{fig:vln-case} presents a navigation example on R2R-CE Val-Unseen. The agent grounds the outside door as the first landmark, turns left toward the stairs, ascends to the upper landing, and follows the instructed hard-left turn toward the workout room. When the direct approach is obstructed, it inspects the surroundings and selects an accessible corridor leading to another entrance. At the final selected state (environment step~64), it confirms arrival inside the workout room and completes the task. The sequence illustrates how landmark grounding, sub-goal execution, local route adjustment, and endpoint verification support instruction following. Figure~\ref{fig:r2r-ceExamples} separately presents navigation comparisons with InternVLA-N1.

\FloatBarrier

\end{document}